\documentclass[12pt]{article}
\usepackage{newtxtext,newtxmath}
\usepackage{graphicx}
\usepackage[letterpaper,margin=1in]{geometry}
\renewenvironment{abstract}
	{\quotation}
	{\endquotation}
\date{}

\makeatletter
\renewcommand{\fnum@figure}{\textbf{Figure \thefigure}}
\renewcommand{\fnum@table}{\textbf{Table \thetable}}
\makeatother
\usepackage{scicite}
\usepackage{url}

\def\scititle{Clustering data by reordering them}
\title{\bfseries \boldmath \scititle}

\author{
	Axel~Descamps$^{1}$, Sélène~Forget$^{1}$, Aliénor~Lahlou$^{2}$, Claire~Lavergne$^{3,4}$, \and Camille~Berthelot$^{3}$, Guillaume~Stirnemann$^{1}$, Rodolphe~Vuilleumier$^{1}$, \and Nicolas~Chéron$^{1\ast}$\and
	\small$^{1}$Chimie Physique et Chimie du Vivant (CPCV), Département de chimie, École Normale Supérieure, \and
	\small PSL University, Sorbonne Université, CNRS, 75005 Paris, France.\and
	\small$^{2}$Sony Computer Science Laboratories, Paris 75005, France.\and
	\small$^{3}$Institut Pasteur, Université Paris Cité, CNRS UMR 3525, INSERM UA12,\and
	\small Comparative Functional Genomics Group, Paris 75015, France.\and
	\small$^{4}$Université Paris Cité, BioSPC, F-75205 Paris, France.\and
	\small$^\ast$Corresponding author. Email: nicolas.cheron@ens.psl.eu\and
}

\begin{document} 
\maketitle

\begin{abstract} \bfseries \boldmath
Grouping elements into families to analyse them separately is a standard analysis procedure in many areas of sciences. We propose herein a new algorithm based on the simple idea that members from a family look like each other, and don't resemble elements foreign to the family. After reordering the data according to the distance between elements, the analysis is automatically performed with easily-understandable parameters. Noise is explicitly taken into account to deal with the variety of problems of a data-driven world. We applied the algorithm to sort biomolecules conformations, gene sequences, cells, images, and experimental conditions.
\end{abstract}

\noindent Grouping data into families to analyse them separately and to identify meaningful patterns is a fundamental task in many branches of science, commonly known as clustering. This can serve different purposes, such as comparing the sizes or dispersions of clusters, examining their properties, or selecting the representative element of the clusters (called centroids) either to reduce analysis complexity or simply to present the data. According to the Cambridge Dictionary, a cluster is ``a group of similar things that are close together, sometimes surrounding something.'' The use of imprecise words, such as ``similar'' and ``close'', highlights the intrinsic challenges of clustering and explains the proliferation of clustering methods. Indeed, while the concept of clustering is intuitively understood, agreeing on a quantitative definition is more challenging. Thus, depending on the question that is asked, the definition that is used, but also the structure of the data, different algorithms can be employed.\\

Most of the existing clustering algorithms require as input a parameter which is impossible to know before-hand: some require the number of clusters ($k$-means\cite{1982:Lloyd:IEEETransactionsonInformationTheory} or Ward\cite{1963:WardJr.:JournaloftheAmericanStatisticalAssociation} methods, e.g.), while other require the density of points (DBSCAN\cite{1996:Ester:ProceedingsoftheSecondInternationalConferenceonKnowledgeDiscoveryandDataMining}, e.g.). For these algorithms, a trial and error approach must be used to compare different values of the parameters. Besides this caveat, in many algorithms all elements are assigned to a cluster and there is no distinction between elements from an actual cluster and elements from noise (which can also contains corrupted data and outliers). Recently, modern methods that can deal with noisy data have been proposed. In the affinity propagation method\cite{2007:Frey:Science}, each point's probability of being a centroid is estimated by evaluating similarities between points; despite some advantages, the algorithm was found to be quite sensitive to the initial probability assigned to each point, which strongly influences the final probability estimation. HDBSCAN\cite{2013:Campello:AdvancesinKnowledgeDiscoveryandDataMining} is an improvement of DBSCAN, which can compare clusters with various densities; however, it can be sensitive to the parameter chosen for estimating the local density. In 2014, the density peaks method\cite{2014:Rodriguez:Science} was proposed: it focuses on identifying centroids by locating high-density zones, and here again the radius chosen to define the local density can significantly influence the algorithm's outcomes (\textit{vide infra}). Thus, all of the existing methods suffer from issues: some are not made to deal with data that contains noise, some require parameters impossible to estimate in advance, and some require a choice to be manually made by the user after a first analysis\cite{2014:Rodriguez:Science}.\\

We propose herein a novel algorithm to clusterize data, which targets several purposes: (\textit{i}) using as few parameters as possible, (\textit{ii}) using only easily understandable parameters, (\textit{iii}) being user-friendly (i.e. not restricted to command-line prompts) and using modern graphical tools, (\textit{iv}) being fast, (\textit{v}) being robust to noisy data. Most of the recently developed clustering methods are density-based models, whereas the current algorithm is somehow closer to the connectivity-based family where the idea is to connect elements that are close to each other. One of the core idea of the proposed algorithm is the reordering of the data according to the distance between elements, and the analysis that is made of the reordered distance matrix. A few methods that reorder the data already exists, OPTICS being the most famous one. However, the reordering made by OPTICS is not done according to the distance and OPTICS depends on a lot of parameters impossible to estimate before-hand. Thus, as will be shown further, OPTICS can fail to properly cluster actual data. Machine learning methods can be highly efficient to separate data into clusters, however not only these methods can fail to properly learn from the data in the case of small datasets, but the sensitivity to parameters can also significantly impact results. We thus envisioned that a simple, efficient, and easy-to-learn algorithm is currently needed. This new algorithm was initially developed to analyse molecular dynamics simulations of biomolecules performed with a technique called replica exchange: the algorithm is thus called YACARE, which stands for ``\textit{Yet Another Clustering Algorithm, but for Replica Exchange}'' (of note, the yacare is a caiman that lives in South America).

\subsection*{The YACARE algorithm}
\begin{figure}
	\centering
	\includegraphics[width=1.0\textwidth]{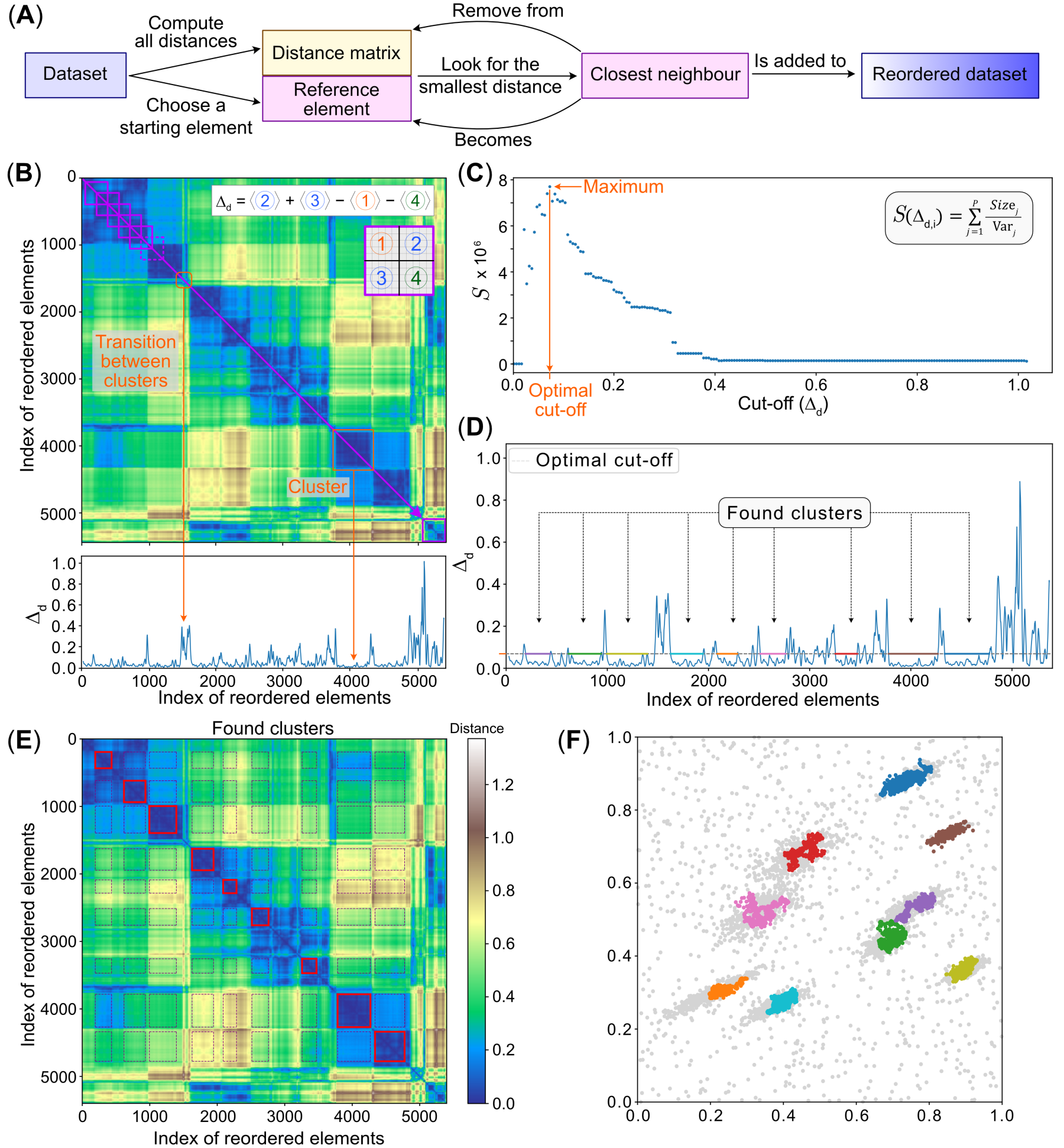}
	\caption{\textbf{Principles of the YACARE algorithm.} (\textbf{A}) Reordering the data. (\textbf{B}) Moving stencil along the diagonal of the reordered matrix and plot of $\Delta_d$ along the diagonal (from the toy dataset). (\textbf{C}) Finding the optimal cut-off. (\textbf{D}, \textbf{E}, \textbf{F}) Final results on the toy dataset, where automatically found clusters are displayed in the plot of $\Delta_d$, in the the distance matrix and in the actual data (prior expansion of clusters).}
	\label{fig:Figure1}
\end{figure}

The core idea of the YACARE algorithm is that elements from a cluster should be close to each other and far from elements of other clusters or from the noise. Thus, the algorithm starts by reordering elements to sequentially identify the closest one neighbour of each element. First, similarity between pairs of elements should be calculated by the user as best fits the nature of the data, which provides a pairwise distance matrix that is used by YACARE. After choosing a starting reference element (which by default is the first one), the element with the smallest distance to the reference one (i.e. the closest neighbour) is identified. This new element becomes the reference structure and the distance matrix is updated to remove the first reference element from the matrix. In addition to becoming the new reference, the index of the closest neighbour is saved to produce afterwards a reordered matrix (see Figure~\ref{fig:Figure1}-(\textbf{A})). Reordering raises the question of the influence of the initial element: several options to try different starting elements are proposed by the algorithm and are discussed in SI.\\

Once different starting elements have been tried, the most optimal one must be found, i.e. the one that will provide the ``best-reordered'' matrix. The optimal path would consist in exploring sequentially the entirety of the elements of one cluster, then jump to the next cluster and explore all of its elements, and so on. In this case, the first off-diagonal in the reordered distance matrix would mainly contain small values of distance, and only a few high distance values for the transition between clusters or from cluster to noise. One could thus compare the first off-diagonal of two reordered matrices to identify the best reordering; this didn't provide significant results, so we decided to additionally take into account the environment of each element. We thus define a stencil consisting of four squares of equal sizes that traverses the diagonal of the reordered matrix. By default, each sub-square contains 0.5\% of the data. For each point along the diagonal, we average the distance in each sub-square and we sum the average of off-diagonal sub-squares (2 and 3 in Figure~\ref{fig:Figure1}-(\textbf{B}), which are small only if the entire stencil is in the same cluster) to which we subtract the average of diagonal sub-squares (1 and 4 in Figure~\ref{fig:Figure1}-(\textbf{B}), which are small when the sub-squares are in a cluster, either the same or two successive ones). This provides a number for each reordered elements, which we call $\Delta_d$ (for \textit{difference along diagonal}): $\Delta_d$ has low values within a cluster, and exhibits maxima when it is at a switching point between two clusters (see Figure~\ref{fig:Figure1}-(\textbf{B})). Since the integral of $\Delta_d$ allows to estimate the number of transitions between clusters for a given reordering, the best starting element for reordering is the one that provides the lowest value for the integral of $\Delta_d$, indicating only a few transitions between clusters.\\

The choice of a cut-off to separate clusters is somehow arbitrary in many clustering algorithms, and it is important to avoid making non-physically-driven choices. Clusters are here identified as zones of data between peaks in the graph of $\Delta_{d}$, i.e. between zones of noise. More explicitly, for a given threshold $\Delta_{d,0}$, a cluster consists of all elements between indices $n$ and $m$, where $n$ is the first index (when moving along the reordered matrix) where $\Delta_d$ drops below $\Delta_{d,0}$, and $m$ is the last index where $\Delta_d$ rises above $\Delta_{d,0}$. The main parameter of the YACARE algorithm is the minimal number of elements that is needed to form a cluster (2.0\% of the dataset size by default), and the chosen cut-off will be the value of $\Delta_{d,0}$ that maximizes a parameter (noted $\mathcal{S}$) that takes into account both the size of the clusters and their homogeneity, and which is defined by default as (see Figure~\ref{fig:Figure1}-(\textbf{C}) and SI for a discussion):
\begin{equation}
    \mathcal{S}(\Delta_{d,i}) = \sum_{j=1}^{P} \frac{Size_j}{Var_{j}}
	\label{eq:OptimalCutOff}
\end{equation}
\noindent where $P$ is the number of clusters for the cut-off $\Delta_{d,i}$, $Size_j$ is the size of cluster $j$ and $Var_{j}$ is the variance of the distance inside the cluster $j$. This procedure provides an objective and non-biased way to determine the optimal cut-off value for separating the clusters. Additional options of the algorithm are discussed in SI, such as merging similar clusters and extending the clusters' sizes. Of note, the YACARE algorithm is thus deterministic, contrary to other methods (such as HDBSCAN or OPTICS) that can lead to a various numbers of clusters when run several times.

\subsection*{Clusterisation of various data}
\begin{figure}
	\centering
	\includegraphics[width=1.0\textwidth]{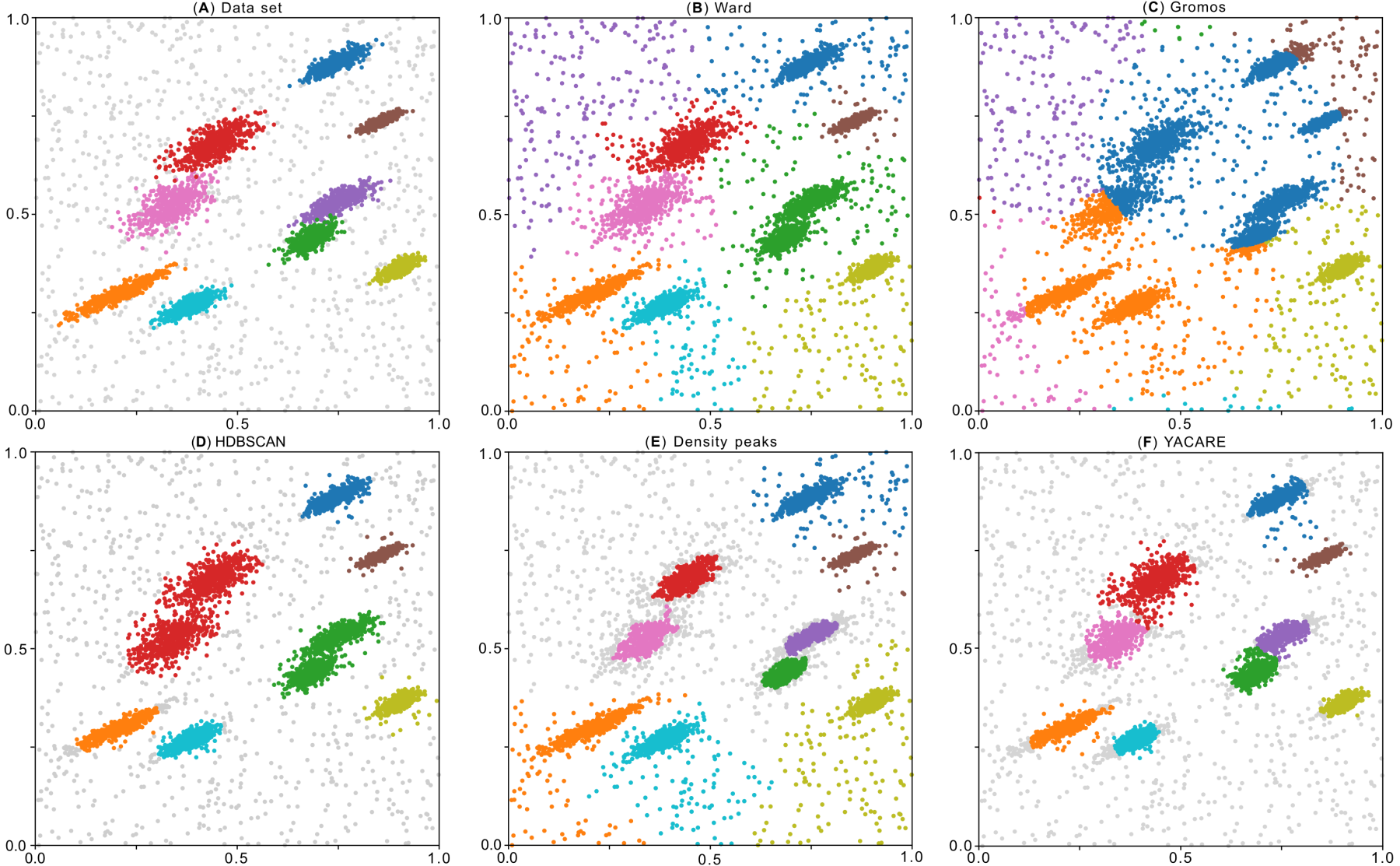}
	\caption{\textbf{Comparison of methods.} (\textbf{A}) Toy dataset with nine sets of 500 points normally-distributed around their centers, and 900 points (20\%) of randomly distributed points. (\textbf{B}) Clustering with the Ward's method, asking for nine clusters, (\textbf{C}) Clustering with the Gromos method, with a cut-off that was chosen to provide nine clusters, (\textbf{D}) Clustering with HDBSCAN, with a minimal cluster size corresping to 3.0\% of the data, (\textbf{E}) Clustering with density peaks, with $\rho_{min}$ and $\delta_{min}$ chosen according to the decision graph, (\textbf{F}) Clustering with the YACARE method with default parameters for merging and expansion of clusters.}
	\label{fig:Figure2}
\end{figure}

The YACARE algorithm was first benchmarked on a toy dataset made with nine sets of 500 points in 2D, normally-distributed around their centers, and with different standard deviations. To these 4500 elements we added 900 (i.e. 20\%) of randomly distributed points (see Figure~\ref{fig:Figure2}-(\textbf{A})). This set was clustered with four existing methods (Ward, Gromos, HDBSCAN and density peaks, Figures~\ref{fig:Figure2}-(\textbf{B}-\textbf{E})) and with YACARE (Figure~\ref{fig:Figure2}-(\textbf{F})); a discussion with four additional methods is proposed in SI. Methods that can't deal with noise fail to properly separate the data (Ward and Gromos), whereas HDBSCAN could properly distinguish noise from data but incorrectly merged two pairs of clusters. The density peaks approach correctly found nine clusters, but for five of them the noise was included in the clusters. On the contrary, YACARE successfully identified nine clusters, with no merging of close clusters, and with no pollution from the noise in the found clusters. The good performance of YACARE was confirmed by the use of four established metrics commonly used to evaluate the efficiency of clustering (adjusted Rand index, adjusted mutual information, V-measure and Fowlkes-Mallows index, see Table~\ref{tab:ComparisonMetrics}): YACARE has the highest score for two metrics, and the second best (very close to the best) for the two remaining metrics.\\

\begin{figure}
	\centering
	\includegraphics[width=1.0\textwidth]{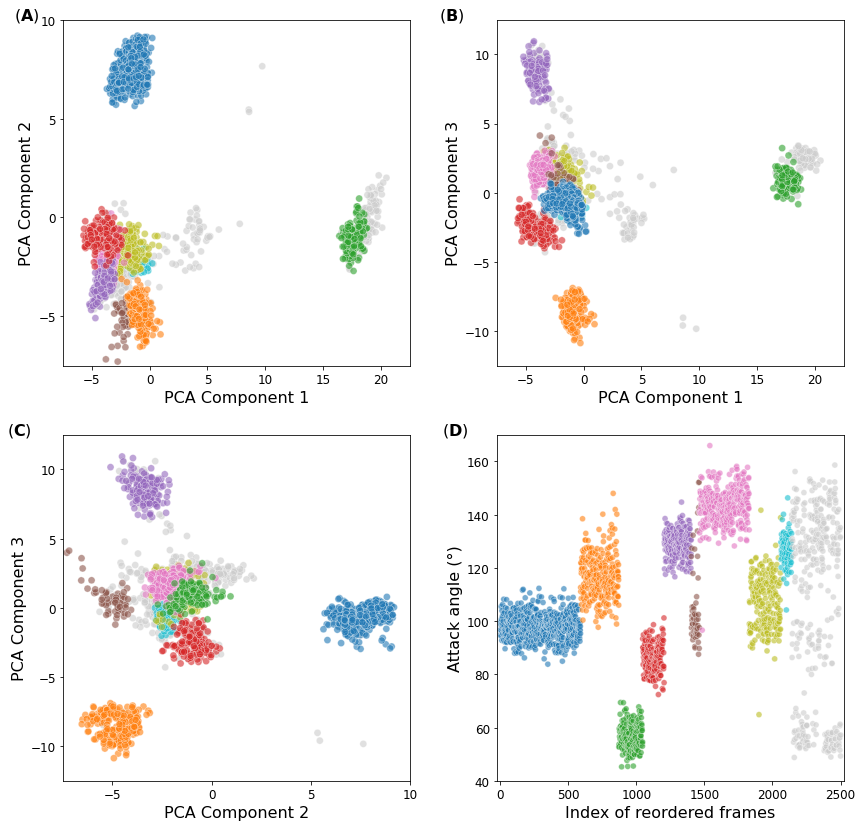}
	\caption{\textbf{Comparison between YACARE and PCA.} (\textbf{A}-\textbf{B}-\textbf{C}) Data points from the RNA structure plotted on two PCA components, and colored according to their clusters. PCA and clustering were performed on the global structure. (\textbf{D}) Local analysis of the attacking angle for the chemical reaction, coming from a 250~ns simulation. The optimal angle is around 140$^{\circ}$ (pink cluster). Data were ordered by YACARE and are colored according to their clusters.}
	\label{fig:Figure3}
\end{figure}

The first scientific question that we addressed is the relevance of the found clusters, using ribonucleic acid (RNA) structures that some of us previously studied\cite{2024:Forget:J.Chem.TheoryComput.}. 1~$\mu$s-long molecular dynamics (MD) simulations with the hamiltonian replica exchange (HREX) method were performed on the RNA hairpin ribozyme, and the last 250~ns were analyzed. The 2500 snapshots were each characterized by 161 contact distances, corresponding to the distances between pairs of nucleobase heavy atoms from non-strand neighbor residues. We only kept distances that remained in close proximity for more than 5\% and less than 95\% of the trajectory, i.e. for pair of atoms that were neither always close nor always far. These data were processed with a principle component analysis (PCA), and were clustered with various algorithms. The overlays of YACARE clusters (with default parameters) on PCA graphs is presented on Figure~\ref{fig:Figure3}, and on Figure~\ref{fig:FigureSI-ARN-OtherMethods} for other clustering methods. We observed that the separation of data into clusters is coherent with what is observed from the separation by PCA: six out of nine clusters from YACARE can be clearly distinguished from the others on at least one PCA plot (blue, orange, green, red, purple and brown). The pink, ocher and cyan clusters stay close and sometimes overlap on the PCA plots, which may raise the question of the separation of these data into three different clusters. The purpose of this biomolecule is to cleave a chemical bound, which happens through the attack of an oxygen onto the phosphorus of a phosphate moiety (P$=$O): for the chemical reaction to occur, the $\widehat{OPO}$ attack angle must be around 140° and the O$-$P distance must be below 3.5\AA. The value of the attack angle observed along the trajectory is presented in Figure~\ref{fig:Figure3}-(\textbf{D}), reordered according to YACARE, and colored according to the clusters found. The pink, ocher and cyan clusters appear here to be well separated, the reactive conformations being grouped in the pink cluster (\#7). Thus, even though the clustering was performed on data that describe the global structure of the RNA ribozyme, a meaningful separation of data at a local scale was observed. Of note, neither OPTICS, HDBSCAN or the density peaks method could capture the existence of the ocher cluster which is always included in the pink one, meaning that the reactive structures found by these algorithms were polluted with irrelevant conformations. The spectral clustering method could capture the existence of the ocher cluster and provides a reactive cluster with relevant conformations, however it wrongly merged the brown and ocher clusters (see SI for a discussion). Thus, only YACARE was able to properly separate the data and to provide useful clusters.\\

The YACARE algorithm was then used to cluster killer cell lectin-like receptor genes (KLR) for 41 mammalian species based on coding sequence similarity. Mammalian immune receptor genes, including KLR, are highly diverse and fast evolving : their evolution into different subfamilies is notoriously difficult to reconstruct and typically requires the use of phylogenetic modeling\cite{2017:Schwartz:Immunogenetics}. With default YACARE parameters, the separation into families of genes was perfectly recovered, with the exception of families with fewer than five elements that were attributed to noise, as well as three genes out of $\sim$300 (see SI for a discussion and Figure~\ref{fig:FigureSI-Genes}-(\textbf{B}) for the confusion matrix between true labels and clustering). In addition to separate genes in clusters, YACARE was also found to be capable of comparing clusters with one another and identifying the closest cluster for each. This allowed to reconstruct, using solely data from YACARE, a phylogenetic tree (see SI for a discussion). On this dataset, HDBSCAN provided comparable results as YACARE (with the exception that one family was split into two clusters), but OPTICS provided too much noise, and density peaks had more confusion between clusters (all confusion matrices are shown in Figure~\ref{fig:FigureSI-Genes-OtherMethods}). To avoid comparing clustering methods solely with confusion matrices --which is prone to a bias in the analysis--, we used established metrics to compare the clustering with labels previously established. For this dataset, the YACARE algorithm provided the highest score for the four metrics presented Table~\ref{tab:ComparisonMetrics} (density peaks being always ranked second or co-first for one of the metrics).\\

Kinetic data were then clustered with YACARE. Single-cell microalgae \textit{Chlamydomonas reinhardtii} were exposed to high light to provoke stress responses, and the evolution of the chlorophyll fluorescence (ChlF) was recorded. This emission reflects the adaptive mechanisms (Non-Photochemical Quenching, or NPQ) put into place at the photosynthetic apparatus level to handle the excess photon energy. This stress response can involve three different NPQ components, and a set of conditioning and selection of mutants allowed to generate populations expressing a single NPQ component \cite{2025:Lahlou:bioRXiv, 2025:NPQScore-data:webpage}, where each population displays a different ChlF pattern when exposed to high-light. A dataset containing the three classes was obtained by segmenting single cells from movies obtained with a fluorescence microscope, yielding 1806 elements. Each element consists of 91 points describing the evolution of ChlF for one cell along time, with a measurement every 20~s during 30~mn. The distance between cells was obtained as the euclidean distance in dimension 91 between the ChlF traces. Using default parameters, YACARE identified seven clusters, five of them being pure or contaminated by a single element; they overall correspond to $\sim$80\% of the data. The two remaining clusters have a low contamination (27 and 12\%) (see Figure~\ref{fig:FigureSI-Kinetics}-(\textbf{B}) for the confusion matrix and SI for a discussion). YACARE was thus found to be efficient for separating cells based solely on a recorded ChlF over time. We compared the use of a minimal cluster size of 2.0\% of the dataset size with a minimal cluster size of two elements, and found that comparable results are obtained (see SI for discussion). Thus, YACARE is robust to the choice of the minimal cluster size. When compared with other algorithms, OPTICS failed here but HDBSCAN and density peaks provided comparable results than YACARE. For this dataset, density peaks provided the best clustering based on the four comparison metrics provided in Table~\ref{tab:ComparisonMetrics}, and YACARE is ranked second in all of them.\\

We next applied YACARE to image recognition, starting with the Olivetti face dataset. This set contains 10 pictures taken from various angles and with different facial expressions for 40 subjects (see Figure~\ref{fig:FigureSI-Olivetti}-(\textbf{A})). This set is known to be challenging due to the low amount of elements in each cluster and to the similitude between subjects. Using the SSIM distance between images, 47 clusters were found (see Figure~\ref{fig:FigureSI-Olivetti}), and with a minimal expansion of clusters we found that 35 clusters contained elements (from two to eight) from a single person. Additionally, 11 actual series of pictures were not splitted in smaller sub-clusters by YACARE. When we forced the assignment of all data into a cluster (i.e. infinite expansion, see SI), 17 clusters stayed pure with data from a single person, and two actual clusters were uncut (see Figures~\ref{fig:FigureSI-Olivetti}-(\textbf{D}-\textbf{E}) for the confusion matrices and SI for a discussion). For this dataset, HDBSCAN and OPTICS failed to provide a satisfactory clustering since the former identified too much clusters (100) and the latter identified too much noise (61\% of the data) (see discussion in SI). When YACARE is compared with density peaks, similar results are obtained: for example, one method identified only pure clusters (30), while the other method identified more clusters (47), not all being pure but with more pure clusters (35). We note however that YACARE identified more faces and had less data in the noise. Based solely on the metrics from Table~\ref{tab:ComparisonMetrics}, density peaks provides higher scores and YACARE is always ranked second.

We then moved to the MNIST database which contains images of handwritten digits from 0 et 9 (see Figure~\ref{fig:FigureSI-MNIST}-(\textbf{A})), using a simple euclidean distance of the pixel values between two images. Even if this problem is significantly more complicated, we still observed that four clusters found by YACARE (out of 29 that were found) were pure and that 13 clusters mainly described a single digit (see Figure~\ref{fig:FigureSI-MNIST}-(\textbf{D}) and SI for a discussion). YACARE could thus be used to separate images in challenging datasets. For this set, all the other methods failed: HDBSCAN identified more than 340 clusters with 70\% of the data in the noise, OPTICS identified 58 clusters with 97\% of data in the noise, and we couldn't find suitable parameters for density peaks that provided more than one cluster (see SI for discussion). As a consequence, YACARE is ranked first when looking at the four comparison metrics from Table~\ref{tab:ComparisonMetrics}.\\

Finally, we turned to clustering chemical reactions. We aimed to determine whether it was possible to identify the original article that described a chemical synthesis based solely on experimental conditions that were reported in the scientific literature. We used the NiCOlit database published in 2022\cite{2022:Schleinitz:J.Am.Chem.Soc.}, which compiles reactions where a nickel-based catalyst activates a C$-$O bond to form a new C$-$C or C$-$N bond. 61 clusters were found by YACARE, and with a minimal expansion of cluster (which leads to 28\% of the data in the noise) 51 clusters were found to be pure (see SI for details on the YACARE parameters). This means that almost all found clusters grouped together data from the same article and that we could successfully separate the chemical reactions. When all data are assigned to a cluster (i.e. infinite expansion), the number of pure clusters decreased to 24, but many of the erroneous assignments were found to be low (see Figure~\ref{fig:FigureSI-NiCOlit}). Noteworthy, several significant confusions between clusters and articles can be explained by the fact that pairs of articles had the same authors (see SI for discussion). For the NiCOlit dataset, HDBSCAN and OPTICS identified too much clusters (170 and 90 respectively). The choice of parameters ($\rho_{min}$ and $\delta_{min}$) from the decision graph for density peaks was not trivial (see Figure~\ref{fig:FigureSI-NiCOlit}-(\textbf{D})), and finally lead to the good amount of clusters but with some confusion in the matrix. When analysing the four comparison metrics from Table~\ref{tab:ComparisonMetrics}, YACARE is always found to provide the highest score for this dataset, sometimes tied with HDBSCAN or OPTICS.\\

We have presented here a novel clustering algorithm that stands out for its simplicity and efficiency. It was shown to be efficient for separating data from various domains of applications, from structures of biomolecules to genes, from images to kinetic data. Using standard clustering performance metrics, YACARE ranked first in 14 instances, outperforming density peaks (9 times), HDBSCAN (4 times), and OPTICS (once). A key strength of YACARE lies in its computational efficiency: by relying solely on pairwise distances without requiring density estimation, it can cluster tens of thousands of elements within a couple of minutes. Moreover, YACARE is highly accessible, with a minimal number of intuitive parameters that can be easily adjusted based on the specific problem at hand. Despite these advantages, we acknowledge that no universal clustering method can exist due to the vast diversity of data structures and research questions. Clustering, much like \textit{haute cuisine}, requires a well-crafted recipe, a skilled chef, and high-quality ingredients. Here, the YACARE algorithm provides the recipe, completed with comprehensive documentation to empower users. The choice of a proper distance metric serves as the essential ingredient that can elevate results from a simple homemade dish to a three-star gourmet experience, and carefully selecting a proper distance is thus of paramount importance for any clustering task and depends on the user.

\begin{table}
\centering
\caption{\textbf{Metrics to compare clustering algorithms.} All the scores are between 0 and 1, higher score meaning better clustering. Best score is in bold. n.a. = not available.}
\label{tab:ComparisonMetrics}
\begin{tabular}{ | c || c || c | c | c | c | }
\hline
Dataset & Clustering     & Adjusted rand & Adjusted mutual & V-measure & Fowlkes-Mallows \\
        & method         & index         & information     &           & index           \\
\hline
         & YACARE	     & \textbf{0.76} & 0.83	         & 0.83	         & \textbf{0.79} \\
Toy      & HDBSCAN	     & 0.72	         & \textbf{0.84} & \textbf{0.84} & 0.76          \\
dataset  & OPTICS 	     & 0.45	         & 0.71	         & 0.71	         & 0.55          \\
         & Density peaks & 0.71	         & 0.82	         & 0.82	         & 0.74          \\
\hline				
       & YACARE	       & \textbf{0.98}	& \textbf{0.97}	& \textbf{0.97}	& \textbf{0.98} \\
Genes  & HDBSCAN	   & 0.94	        & 0.96	        & 0.96	        & 0.95          \\
       & OPTICS	       & 0.06	        & 0.37	        & 0.46        	& 0.25          \\
       & Density peaks & 0.95	        & 0.96	        & \textbf{0.97} & 0.95           \\
\hline
         & YACARE        & 0.77	         & 0.72	         & 0.72	         & 0.86          \\
Kinetics & HDBSCAN	     & 0.73	         & 0.47        	 & 0.51	         & 0.83          \\
         & OPTICS    	 & 0.01	         & 0.08       	 & 0.10	         & 0.59          \\
         & Density peaks & \textbf{0.87} & \textbf{0.81} & \textbf{0.81} & \textbf{0.92} \\
\hline
         & YACARE	     & 0.42	         & 0.59          & 0.76	         & 0.44          \\
Olivetti & HDBSCAN	     & 0.21	         & 0.55          & 0.79 	     & 0.24          \\
         & OPTICS 	     & 0.04	         & 0.39	         & 0.56	         & 0.18          \\
         & Density peaks & \textbf{0.49} & \textbf{0.71} & \textbf{0.82} & \textbf{0.51} \\
\hline
       & YACARE  	    & \textbf{0.25}	& \textbf{0.46}	& \textbf{0.47} & \textbf{0.32} \\
MNIST  & HDBSCAN	    & 0.04	        & 0.28	        & 0.33	        & 0.28          \\
       & OPTICS 	    & 0.00	        & 0.01	        & 0.02	        & 0.31          \\
       & Density peaks	& n.a.          & n.a.	        & n.a.	        & n.a.          \\
\hline
        & YACARE 	    & \textbf{0.51}	& \textbf{0.77}	& \textbf{0.81}	& \textbf{0.54} \\
NiCOlit & HDBSCAN 	    & 0.45        	& 0.74	        & \textbf{0.81}	& \textbf{0.54} \\
        & OPTICS 	    & 0.45	        & 0.75	        & \textbf{0.81}	& 0.50          \\
        & Density peaks & 0.45	        & 0.75	        & 0.78	        & 0.48           \\
\hline
\end{tabular}
\end{table}

\clearpage
\bibliography{YACARE}
\bibliographystyle{sciencemag}

\section*{Acknowledgments}
The authors thank Jérôme Hénin and Elise Duboué-Dijon for fruitful scientific discussions, as well as Julie Puyo and Julia Attard for beta-testing the algorithm. 

\paragraph*{Funding:}
This work was granted access to the HPC resources of CINES and IDRIS for N.C. under the allocation 2023-077156 made by GENCI. The research leading to these results has received funding from the European Research Council under the European Union's Eighth Framework Program (H2020/2014-2020)/ERC grant agreement no. 757111 (G.S.).

\paragraph*{Author contributions:}
Conceptualization: A.D., R.V., N.C. Data curation: S.F. and G.S. (RNA), C.L. and C.B. (genes), A.L. (kinetic). Formal analysis: A.D. and N.C. Investigation: A.D. and N.C. Methodology: A.D. and N.C. Project administration: N.C. and R.V. Software: A.D. and N.C. Supervision: N.C. Validation: N.C. Writing - original draft: A.D. and N.C. Writing - review \& editing: all authors.

\paragraph*{Competing interests:}
There are no competing interests to declare.

\paragraph*{Data and materials availability:}
The \textsc{Python} code of YACARE is made available with the current article. Th RNA simulation used in this article was published recently \cite{2024:Forget:J.Chem.TheoryComput.}. Data used for the genes were obtained as explained below. Kinetics of fluorescence are publicly available \cite{2025:Lahlou:bioRXiv, 2025:NPQScore-data:webpage}. The Olivetti dataset was obtained through \textsc{Scikit-learn}. The MNIST dataset was downloaded at the adress https://pjreddie.com/projects/mnist-in-csv/. The NiCOlit dataset for chemical reactions is coming from Ref. \cite{2022:Schleinitz:J.Am.Chem.Soc., 2025:NiCOlit:webpage}.

\subsection*{Supplementary materials}
Materials and Methods\\
Supplementary Text\\
User Manual\\
Table S1\\
Figures S1 to S18\\
References \textit{(13-\arabic{enumiv})}\\ 
Movies S1 and S2

\newpage

\renewcommand{\thefigure}{S\arabic{figure}}
\renewcommand{\thetable}{S\arabic{table}}
\renewcommand{\theequation}{S\arabic{equation}}
\renewcommand{\thepage}{S\arabic{page}}
\setcounter{figure}{0}
\setcounter{table}{0}
\setcounter{equation}{0}
\setcounter{page}{1} 

\begin{center}
\section*{Supplementary Materials for\\ \scititle}
Axel~Descamps, Sélène~Forget, Aliénor~Lahlou, Claire~Lavergne, \\
Camille~Berthelot, Guillaume~Stirnemann, Rodolphe~Vuilleumier, Nicolas~Chéron$^{\ast}$ \\	
\small$^\ast$Corresponding author. Email: nicolas.cheron@ens.psl.eu
\end{center}

\subsubsection*{This PDF file includes:}
Materials and Methods\\
Supplementary Text\\
User Manual\\
Table S1\\
Figures S1 to S18

\newpage
\subsection*{Materials and Methods}
YACARE is written in \textsc{Python} and is made of $\sim$2,500 lines of code. It only uses standard package: \textsc{Numpy}, \textsc{Matplotlib} and \textsc{Scikit-learn}. When one wants to draw a confusion matrix, \textsc{Seaborn} is needed. YACARE will be made available on https://pypi.org/ once the article is accepted. Thus, YACARE will be installable with ``pip install yacare\_cluster''. It is possible to use YACARE solely with a command-line interface, which can be useful in the case of large datasets when one wants to parallelize the search of the best starting point for reordering. However, it is advised to use the provided \textsc{Jupyter} notebook to use YACARE, since it allows to seamlessly analyse the output images (see user manual below). We provide below additional details regarding the algorithm.

\subsubsection*{Reordering the data}
After the first reordering is done (starting from the first element of the data), a choice is proposed to the user to try other starting points: he/she can (1) start from the centroids of clusters that were identified during the first reordering, (2) try a given number of randomly chosen elements, (3) try a given number of evenly separated elements, (4) try all the elements, or (5) specify the index that provides the best reordering if it was determined before. The choice is directly proposed in the provided \textsc{Jupyter} notebook. For up to $\sim$2,000 elements in the dataset, reordering from all the elements takes only a few minutes on a laptop computer. In all the examples presented here, this is the strategy that we have used and is the recommanded one.\\

When data are reordered with a starting element which is not the optimal one, the worst that can happen is the splitting of a cluster in several smaller ones. This will happen if two clusters A and B are both close to each other and dispersed: during the reordering, data from half of cluster A could be processed, and when at the border of cluster A the closest neighbour may be from cluster B and not from the other half of cluster A. If such a case occurs, cluster A will be splitted in two (A-1 and A-2), and both clusters will have the same intrinsic properties. Should this case arise, this will be seen in the provided figures such as Figures~\ref{fig:FigureSI-Tools}-(\textbf{B}-\textbf{C}). The option to merge clusters (describe below) will then automatically concatenate clusters A-1 and A-2.\\

We present in Figure~\ref{fig:FigureSI-CompareReordering} the distance matrix from the toy dataset reordered according to six different elements: the first one, the ones that provide the lowest and highest values for the integral of $\Delta_d$, and three random elements. Differences appear between these six matrices, but comparing them solely with the eye would be difficult and prone to errors.

\subsubsection*{Choice of cut-off}
To identify the cut-off, we rely on the analysis of $\Delta_d$. We present in Figure~\ref{fig:FigureSI-CompareDistances} the overlap between $\Delta_d$ and the values in the first off-diagonal (i.e. the sequential distance with the first neighbour of each element). One can clearly see that the first off-diagonal is highly noisy. Performing a running average of the off-diagonal smoothens the signal as expected, but provides a different curve than $\Delta_d$ since the heights of the peaks are different. This explains why we decided to take into account the environment of each element with the stencil and not information from only the closest neighbour.\\

To find the optimal cut-off, the interval between the minimum and the maximum of $\Delta_d$ is evenly splitted (by default in 200), and each value of $\Delta_{d,i}$ is tested. When using the function defined in main text for $\mathcal{S}$, for some rare cases, the plot of $\mathcal{S}$ with respect to the cut-off values didn't display a maximum but instead had the shape of a sigmoid. In these cases, the size of the clusters dominated the sum of ratios in the formula of $\mathcal{S}$. Such cases can easily be identified by eye, and we propose in such cases to instead sum the inverse of variances for all clusters without taking into account the size of the clusters.\\

Following the first identification of clusters, the user has the choice to increase or decrease the proposed cut-off to either separate a large cluster in two, to merge two clusters that seem to be close, or to capture an otherwise-missed cluster. This is done by directly providing another value for $\Delta_{d,0}$ that will be used as a cut-off. The decision can be guided by the visual analysis of the graph of $\Delta_d$ and of the distance matrix. We found that increasing the cut-off solely to merge two clusters that seemed to be close from the analysis of $\Delta_d$ is usually a bad idea and didn't change the proposed cut-off.

\subsubsection*{Merging and expanding clusters}
A graphical analysis of all clusters and off-diagonal zones is provided in the notebook as in Figure~\ref{fig:FigureSI-Tools}-(\textbf{B}-\textbf{C}), where one can (\textit{i}) compare the size of all clusters, and the average distance and deviation within each cluster, as well as within all off-diagonal zones of the reordered distance matrix, (\textit{ii}) see the position of clusters within the reordered matrix, where clusters are identified as red squares. Following this visual analysis, the option to merge several clusters is proposed (we point out that this will lead to different results than changing the cut-off value). The problem that was mentioned above that could occur with a non-optimal reordering starting point can easily be solved at this stage, since the off-diagonal zones will appear as having a low distance between A-1 and A-2. To decide to merge two clusters A and B, we compare the mean distance within the out-of-diagonal zone (in dashed purple) between clusters A and B to a threshold value that is defined as $\mu_A + \alpha \cdot \sigma_A$ where $\mu_A$ and $\sigma_A$ are the mean and standard deviation from cluster A (i.e. from the red square in the diagonal) and $\alpha$ is a parameter (1.0 by default) that can be chosen by the user. Cluster B will be merged with cluster A if the out-of-diagonal mean distance is below the threshold of distances within A.\\

Since the distance matrix is symmetric, the out-of-diagonal zones between A and B on one side, and between B and A on the other side, are equivalents. However, the threshold value depends on values from the clusters, and is thus different for clusters A and B. Thus, it may be acceptable to merge A with B, but not B with A. As a consequence, the order that is used for merging the clusters matters, and the user can choose to start from (\textit{i}) the smallest cluster, (\textit{ii}) the largest cluster, (\textit{iii}) the cluster with the highest number of neighbours, (\textit{iv}) the cluster with the fewest number of neighbours. By default, we propose to start with the cluster that has the highest number of neighbours (option \#3). A list for merging clusters is automatically proposed, and this list can then be manually modified following the visual analysis of the matrices to match the user's needs.\\

As can be seen in Figure~\ref{fig:FigureSI-Tools}-(\textbf{C}), the identified clusters are small compared to the area where the distance appears to be low. This is a behaviour that was purposely pursued during the development of the code, since it allows to detect smaller but purer clusters. We preferred to be quite strict in the size of clusters, and then propose the option to merge clusters as seen above. The last option that is proposed to the user is to expand a cluster to check in the noise if some elements should have been assigned to clusters. Indeed, since $\Delta_d$ cannot be computed for the first and last elements of the reordered matrix due to the use of a stencil, the expansion of clusters allows to rescue these data. To perform the expansion, for each element that was found in the noise, we first look to which cluster it is the closest to, and then check if it is at a distance lower than $\mu + \beta \cdot \sigma$ to the center of the closest cluster. Here, $\mu$ and $\sigma$ are the mean distance and the standard deviation inside the closest cluster, and $\beta$ is a parameter (1.0 by default) that can be chosen by the user. Its value will depend on the question being addressed: the need is not the same if one wants to compare the size of two clusters, or if one wants pure clusters for example to extract representative structures. If the data is not supposed to contain noisy elements, the user can add the option \verb"keep_no_noise=True" which is equivalent as having $\beta$=$\infty$.With the toy dataset presented in main text, 47\% of the data where initially found in the noise (see Figure~\ref{fig:Figure1}-(\textbf{G})), which decreased to 29\% after expanding the clusters with $\beta$=1.0 (see Figure~\ref{fig:Figure2}-(\textbf{F})). With $\beta=2.0/3.0/5.0/10.0$, the amount of data in the noise decreased to $20/14/10/4\%$. As a recall, the toy dataset is made of 4500 points plus 20\% of random noise, thus noise accounts for 17\% of the data. However, since noise is randomly distributed and can overlap the actual data, only 14\% of the data can theoretically be identified as noise.

\newpage
\subsection*{Supplementary Text}
In the following, we provide additional details regarding the use of YACARE with several applications. Results will be mainly displayed in the form of confusion matrices, which are tables where each line describes an already-found cluster (so-called ``truth'') and each column describes a cluster found by  an algorithm. The value in line $i$ and column $j$ indicates the amount of elements in the actual cluster $i$ that were found in the cluster $j$ by the algorithm. A perfect tables would be squared, with non-zero values only along the diagonal. Since the purpose of YACARE is to identify clusters as pure as possible (i.e. with the fewest mix between actual clusters), we are expecting more column than lines. In such a case, a balance has to be found in the number of clusters since one doesn't want to obtain a non-realistic amount of clusters which would all be pure but really small. A ``pure'' cluster is defined as a cluster found by YACARE with elements coming from a single actual cluster, i.e. a column with a single non-zero value. When clusters are found by YACARE, there is no reason why cluster \#1 (for example) should match the actual cluster \#1. Thus, to ease the analysis, the columns of the confusion matrices were automatically reordered to provide images more easy to analyse. When we compare different values of $\beta$, the same reordering of the columns were done so that column \textit{j} with a given value of $\beta$ matches column \textit{j} with another value of $\beta$.

\subsubsection*{Clustering of the toy dataset with other methods}
In main text, we compared YACARE with four methods for clustering the toy dataset. We present in Figure~\ref{fig:FigureSI-OtherMethods} the clustering with three additional methods: \textit{k}-means\cite{1982:Lloyd:IEEETransactionsonInformationTheory}, OPTICS (Ordering Points To Identify the Clustering Structure)\cite{2011:Kriegel:WIREsDataMiningandKnowledgeDiscovery} and spectral clustering\cite{2000:Shi:IEEETransactionsonPatternAnalysisandMachineIntelligence, 2001:Ng:AdvancesinNeuralInformationProcessingSystems}. Affinity propagation\cite{2007:Frey:Science} was tried, but we couldn't find a value for the ``preference'' parameter that proposed nine clusters and the default value proposed 1696 clusters, denoting an issue with this method.

\textit{k}-means wrongfully merged two clusters (Figure~\ref{fig:FigureSI-OtherMethods}-(\textbf{G}), in green) and proposed a cluster made solely with data from the noise (in purple). The clusters identified by OPTICS were clean and properly separated from the noise, but only seven clusters were found (two actual clusters were attributed to the noise). Spectral clustering identified the nine clusters, but since this approach can't deal with the noise, clusters were found to be too wide. Moreover, the transition between close clusters (green and purple for example, or orange and pink) were not properly assigned.

\subsubsection*{Clustering of the RNA ribozyme simulations}
We analyzed molecular dynamics (MD) simulations coming from \cite{2024:Forget:J.Chem.TheoryComput.}. Briefly, a RNA ribozyme built from the precatalytic state structure of the minimal hairpin ribozyme (PDB ID 2OUE \cite{2006:Salter:Biochemistry}) was solvated in a 0.2M KCl solution (see Figure~\ref{fig:FigureSI-ARN-Structure}). Simulations were performed with Gromacs2019.4 \cite{Hess:JCTC:2008:Gromacs, Pronk:Bioinfo:2013, Abraham:SoftwareX:2015} with a 2~fs timestep, the RNA being described with the AMBER ff99 force field \cite{1999:CheathamIII:JournalofBiomolecularStructureandDynamics}, corrected by the Barcelona $\alpha$/$\gamma$ \cite{2007:Perez:BiophysicalJournal} and the Olomouc $\chi_{OL3}$ and $\epsilon\zeta_{OL1}$ reparametrizations \cite{2011:Zgarbova:J.Chem.TheoryComput., 2013:Zgarbova:J.Chem.TheoryComput.}; the TIP3P force field was used to describe water \cite{1981:Jorgensen:J.Am.Chem.Soc.}. After an equilibration step in which the catalytic site was constrained into a specific conformation \cite{2024:Forget:J.Chem.TheoryComput.}, a 1$\mu$s-long simulation of hamiltonian replica exchange (REST2 framework \cite{2011:Wang:J.Phys.Chem.B} as implemented in Plumed \cite{2009:Bonomi:ComputerPhysicsCommunications, 2014:Tribello:ComputerPhysicsCommunications, 2019:Bonomi:NatMethods}) was performed with 24 replicas and a range of temperature of 300-450K. Simulations were performed in the NPT ensemble with the velocity-rescale thermostat \cite{Bussi:JCP:2007} and the Parrinello-Rahman barostat \cite{Parrinello:JAP:1981}. Electrostatic interactions were described with the particle-mesh Ewald method \cite{Darden:JCP:1993}, with a Verlet cutoff scheme of 10.0\AA\ for Lennard-Jones interactions. Bonds containing an hydrogen were restrained with the LINCS algorithm \cite{Hess:JCC:1997, Hess:JCTC:2008:LINCS}.\\

Visually analysing raw trajectories from HREX simulations is nearly impossible, due to the frequent jump and jiggling of atoms which are a consequence of the regular exchange of conformations (see Movie S1). Since in replica exchange trajectories the sequential order of frames is not relevant, reordering the frames is legit and should lead to a reduction of the observed abrupt discontinuities in the time evolution of observables (of note, it is this statement that initially lead to the development of the algorithm). Using YACARE, one can produce a reordered trajectory that present smooth transitions between structures which can be of significant help during the analysis. Such a trajectory is presented in Movie S2 and is much easier to analyse than raw data from MD seen in Movie S1.\\

To describe the RNA structures, 161 contact distances were chosen and extracted for structures every 100~ps along the last 250~ns of the simulation (i.e. the structure was equilibrated during 750~ns). We chose distances between the heavy atoms of pairs of nucleobases from non-strand neighbor residues. Including in this set pairs of nucleobases that are never formed or always formed would pollute the data, we thus only included pairs that remained in close proximity for more than 5\% and less than 95\% of the trajectory. After normalisation, these data were processed with a principle component analysis (PCA) method to reduce the dimensionality of the problem. 25 components accounted for more than 1\% of the description of the data, the first three accounting for 30, 14 and 12\% i.e. a total of 56\%. The next three accounted for 10, 6 and 4\% of the data.\\

We present in Figure~\ref{fig:FigureSI-ARN-OtherMethods} the projection of the clusters found by various methods on the PCA plots, similarly as Figure~\ref{fig:Figure3} of main text. The blue, orange, green and purple clusters can be clearly separated from the other clusters in at least one of the plots for all the presented methods (spectral clustering, HDBSCAN, OPTICS, density peaks). When it exists, the brown cluster can be separated from the others in the last plot. The separation is less obvious for the red cluster. For HDBSCAN, OPTICS and density peaks, the pink and ocher clusters that were found separate by YACARE are merged into a single one (presented here as pink); this means that the conformations with the correct attack angle (pink cluster \#7 from YACARE) are polluted with irrelevant conformations from the ocher cluster with HDBSCAN, OPTICS and density peaks. The cyan cluster can be found with these three methods, but sometimes with a very small size (only three elements in density peaks). The spectral clustering method separated the pink and ocher clusters, but included the brown cluster in the ocher one: this is acceptable when one focuses on the attack angle since brown and ocher clusters represent conformations with roughly the same value of the angle, but is not acceptable when one focuses on the second component of PCA (see third panel of Figure~\ref{fig:FigureSI-ARN-OtherMethods}). We also note that an extra cluster (presented as light green) was found by spectral clustering. The affinity propagation method was tried on these data, but it provided 76 clusters and we couldn't identify parameters that lead to useful results. Thus, for the RNA simulation, only YACARE provided a separation of data that matches the PCA analysis and the local analysis of the angle.\\

We present in Figure~\ref{fig:FigureSI-ARN-AnglesDistances} the two-dimensional densities of the relevant angle and distance for the 2500 snapshots of the RNA ribozyme. Meaningful structures for the reactivity should be located inside the blue rectangle, and it appears clearly that only cluster \#7 (presented in pink in main text) can describe a reactive conformation.

\subsubsection*{Clustering of genes}
We curated the annotated killer cell lectin-like receptor (KLR) genes for 41 species, with available reference genomes through Ensembl (v. 111), Ensembl Rapid Release, the National Center for Biotechnology Information (NCBI), and the Bat1K Consortium. From these 41 genomes, we identified the annotated KLR genes by using BioMart for the Ensembl genomes (attributes: gene stable ID, gene name), by filtering the gene gtf files for Ensembl Rapid Release and NCBI genomes, and by filtering the TOGA ortholog classification file for the Bat1K genomes. We also searched using known synomyns of KLR genes: Ly49 and NKG2. We obtained the coding sequence (CDS) of each gene, using the longest transcript. Low quality sequences were removed when flagged as such by the reference genome source. Using the C2A.A2C (v. 1.40) tool, we translated the CDS into amino acids. The total number of sequences was 302. We then aligned the sequences using MAFFT (v. 7.50) with the \verb"--globalpair --maxiterate 1000" options. To calculate the distance matrix of the alignment, we used the DistanceMatrix() function with default parameters from the R package DECIPHER (v. 2.28.0). The phylogenetic tree was built with IQ-TREE (v. 2.2.2.2), using the following parameters \verb"-m Q.mammal+I+G4 -alrt 1000 -B 1000". The best-fit model was selected with the \verb"-m TESTONLY" option on a previous run. SH-like approximate likelihood ratio test (SH-aLRT) and ultrafast bootstrap were performed to gauge the robustness of the tree. The tree was visualized in FigTree (v1.4.4), with SH-aLRT and bootstrap values displayed at each node.\\

The clustering of genes with YACARE was performed with default parameters (stencil size = 1.0\%, minimal size of clusters = 2.0\%, $\beta$ = 1.0 to expand the clusters) and nine clusters were found. To merge clusters, we used $\alpha$ = 1.0, but no merging was proposed. The confusion matrix between previously identified clusters and results from YACARE is displayed in Figure~\ref{fig:FigureSI-Genes}-(\textbf{B}) and shows a near perfect agreement. In addition to grouping genes in families, the YACARE algorithm was also found to be able to capture similarities between families. Indeed, by comparing the mean distances in the out-of-diagonal zones, one can identify the cluster B that is the closest to a given cluster A. There is no reason why this should be symmetric, and the closest cluster to cluster B may a third cluster C. However, if B is the closest to A and A is the closest to B, this means that clusters A and B form a sub-family of clusters, which for evolution means that they diverged recently. In Figure~\ref{fig:FigureSI-Genes}-(\textbf{A}), the gray arrows indicates the closest cluster from each cluster. Cluster \#13 is found by YACARE to be the closest to cluster \#4, and \textit{vice versa}, they should thus be grouped together. The closest cluster to cluster \#3 is cluster \#4, thus cluster \#3 can be added to the family of clusters \#4/\#13, but with a divergence which occurred longer ago. The same analysis can be performed for clusters \#6 and \#7 first, and then cluster \#2. The closest cluster from cluster \#9 is cluster \#8, thus these two should be in the same branch of evolution. Finally, the closest cluster from clusters \#1 and \#8 is cluster \#2 which again is the indication of an early divergence in evolution. All these results from YACARE match the phylogenetic tree obtained previously.\\

Genes that were found in the noise are labelled with red stars in Figure~\ref{fig:FigureSI-Genes}-(\textbf{A}). With default parameters, YACARE was not able to capture small gene families, such as clusters \#5/\#10/\#11 which contains only 3, 2 and 5 genes respectively. However, if the expansion of cluster is made with $\beta$ = 2.0 (instead of 1.0 by default), the five genes from cluster \#11 are assigned to cluster \#3 which is consistent with the phylogenetic tree. Beside the clusters \#5/\#10/\#11 and the two isolated genes, only three genes in clusters were found by YACARE in the noise which may be a sign that these genes have started to evolve and are currently diverging from their current cluster.\\

We present in Figure~\ref{fig:FigureSI-Genes-OtherMethods} the clustering of the genes data with HDBSCAN, OPTICS, and density peaks. Results from HDBSCAN are similar to those from YACARE, with the same amount of noise; the sole main difference is the fact that cluster \#3 was split in two in HDBSCAN (and three genes were attributed to noise). With OPTICS, 60\% of the data were attributed to noise with default parameters; several parameters can be changed to modify the amount of noise, but this may lead to a cherry-picking situation where the user adapts the set of OPTICS parameters for each initial dataset, and thus we didn't try to optimize the OPTICS parameters. With density peaks, no noise was identified and all genes were attributed to one of the nine found clusters; as a consequence, only five clusters were found to be pure and the four remaining clusters showed signs of confusion with other clusters. Thus, for this example, YACARE and HDBSCAN provided similar results, both being superior to OPTICS and density peaks. 

\subsubsection*{Clustering of fluorescence kinetic data}
In the dataset of fluorescence kinetic data, each fluorescence trace characterises the stress response of a single-cell alga (\textit{Chlamydomonas reinhardtii}) induced by exposure to high-light for 15~minutes, followed by 15~minutes of relaxation in the dark. In order to record the fluorescence across the whole experiment, even in the dark conditions, a strong light pulse was applied for 200~ms every 20~seconds to read the state of the system. We recorded only the chlorophyll fluorescence (ChlF) response corresponding to these strong pulses, which reflects the adaptive mechanisms used by the organism to release the excess absorbed energy. This so-called non-photochemical quenching (or NPQ) can occur following three pathways (called qE, qT and qI), and a set of conditioning and a selection of mutants allowed to generate populations expressing a single NPQ component and elementary ChlF traces (labelled Pop-qE, Pop-qT and Pop-qI). During the experiments, the ChlF response was monitored along time under a microscope with hundreds of algae in the field of view. The used dataset is available online \cite{2025:Lahlou:bioRXiv, 2025:NPQScore-data:webpage}, and the population expressing no NPQ component was not considered here because we were interested in signatures that exhibit a response. The curated dataset consisted of 1806 cells. Each class of cells displayed a different ChlF pattern reflecting the NPQ component. Following the experiments, a vector of 91 values was obtained for each alga (see Figure~\ref{fig:FigureSI-Kinetics}-(\textbf{A})) and the similarity between cells was obtained as the euclidean distance in dimension 91 after standardization of the data (even though in this case, similar results were obtained without standardization since all the 91 features are equivalent).\\

When we used the default stencil size of 1.0\%, the graph of $\Delta_d$ appeared noisy (see Figure~\ref{fig:FigureSI-Kinetics}-(\textbf{C})). We thus increased the stencil size to 2.0\% (see Figure~\ref{fig:FigureSI-Kinetics}-(\textbf{D})) and used the default minimal cluster size of 2.0\%. 12 clusters were found and are displayed Figure~\ref{fig:FigureSI-Kinetics}-(\textbf{E}). We used the automatic option to merge clusters with $\alpha$ = 1.0 (i.e. default parameter), starting with the cluster with the highest number of neighbors; clusters \#1/\#2/\#3/\#4/\#7 were proposed to be merged, as well as clusters \#5 and \#6, which resulted in seven final clusters. The color of clusters in Figure~\ref{fig:FigureSI-Kinetics}-(\textbf{E}) that were merged match the ones in Figure~\ref{fig:FigureSI-Kinetics}-(\textbf{F}). We then expanded clusters with $\beta$ = $\infty$ in order to capture all data. This provided the matrix and the clusters in Figure~\ref{fig:FigureSI-Kinetics}-(\textbf{F}), which was then used to compute the confusion matrix where the columns were shuffled to ease the readability. Four clusters (\#1, \#5, \#6 and \#7) were found to be pure and to describe respectively mutant 1, 2, 3 and 3. Cluster \#3 was found to be contaminated by a single element and to describe mutant 2. Clusters \#2 and \#4 mainly (at 73 and 88\%) describe one mutant (respectively mutants 1 and 2). From the visual inspection of the final clusters in Figure~\ref{fig:FigureSI-Kinetics}-(\textbf{F}), the first two clusters (orange and purple) seem similar enough to be merged, as well as the last three (pink, ochre and cyan). Columns in the confusion matrix are shuffled to increase the readability, and the first two clusters in Figure~\ref{fig:FigureSI-Kinetics}-(\textbf{F}) correspond to clusters 6/7 in the confusion matrix, whereas the last three clusters correspond to clusters 3/4/5 in the confusion matrix. Thus, using results from the visual inspection would have lead to only four clusters, one of them describing mutant 2 at 94\% and another one describing mutant 3 at 100\%.\\

Instead of using a minimal cluster size of 2.0\%, another way to use YACARE is to identify a lot of small clusters and let the algorithm merge them. If one proposes a cluster size that is too small, the program will automatically increases the threshold so that clusters are made of at least two elements. When applied to the clustering of kinetic data, the minimal cluster size proposed by the algorithm was 0.12\%. This resulted in 45 found clusters as shown in Figures~\ref{fig:FigureSI-Kinetics-SmallerSize}-(\textbf{A}-\textbf{B}). When using the default parameters for merging ($\alpha$=1.0), 11 clusters were proposed. However, it makes sense to increase this threshold since more clusters were produced. With $\alpha$=2.0 seven clusters were found, and with $\alpha$=3.0 four clusters were found (see confusion matrices displayed Figures~\ref{fig:FigureSI-Kinetics-SmallerSize}-(\textbf{C}-\textbf{D}) (in all cases, the expansion of clusters was made with $\beta$=$\infty$). We observe that with $\alpha$=2.0, six out of seven clusters were pure or contained a single outlier, whereas with $\alpha$=3.0 two clusters were pure and a third one contained 63\% of data from a single mutant. Thus, the YACARE algorithm is robust to the choice of the minimal cluster size as well as to the choice of the merging of clusters.\\

We present in Figure~\ref{fig:FigureSI-Kinetics-OtherMethods} the clustering of the genes data with HDBSCAN, OPTICS and density peaks. With a minimal cluster size of 2.0\% of the data, HDBSCAN provided good results with three found clusters, two being pure and the third one representing mutant 2 by 84\%; 18\% of noise was found. OPTICS completely failed here and found only one cluster and noise. Density peaks provided acceptable results with four clusters, however we note that no cluster that described purely mutant \#1 was identified. When we changed the minimal cluster size to 0.12\% (as we did with YACARE), HDBSCAN failed by finding 168 clusters, with no possibility to directly merge them (we note however that all these clusters were pure); in that case, 23\% of noise was found. OPTICS found 19 clusters (all pure) but they corresponded to only 6\% of the data and 94\% of noise was found. Similarly to the discussion with genes, the amount of noise could be decreased, but this would require a manual choice to be made by the user that depends on the dataset. Thus, for this example, YACARE, HDBSCAN and density peaks provided comparable results but for different reasons: with default parameters, YACARE has more clusters than expected (seven instead of three) but finds clusters that can purely describe each mutant and includes a control of the amount of noise, HDBSCAN finds the correct number of clusters but with less control on the noise and no pure cluster \#2, and density peaks finds four clusters and finds no noise but with no pure cluster \#1.

\subsubsection*{Clustering of the Olivetti face dataset}
Images from the Olivetti database were fetched from \textsc{Scikit-learn} (see Figure~\ref{fig:FigureSI-Olivetti}-(\textbf{A})), and the distance between two images were computed with the structural similarity index measure (SSIM) method which provides a number indicating the amount of similarity between images. The initial plot of $\Delta_d$ appeared noisy with the default stencil size of 1.0\% (Figure~\ref{fig:FigureSI-Olivetti}-(\textbf{B})), we thus increased it to 2.0\%. Since we wanted to capture a large amount of cluster, we used the minimal cluster size proposed by the algorithm, which is 0.51\% (which corresponds to a minimum of two elements per cluster), which lead to 47 clusters (Figure~\ref{fig:FigureSI-Olivetti}-(\textbf{C})). We didn't try to merge clusters, and then compared different values of $\beta$ for expanding the clusters. Results are presented in Table \ref{tab:SI-OlivettiData} and confusion matrices with $\beta$=0.0 and $\beta$=$\infty$ are displayed Figures~\ref{fig:FigureSI-Olivetti}-(\textbf{D}-\textbf{E}). Depending on the question that is asked (for example having pure clusters, or assigning all data to a cluster), one will have to choose a different value of $\beta$. With $\beta$=0.0, one will prefer to have pure clusters. In this case, most of the found clusters (35) didn't mix faces and contained elements (from two to eight) from a single person. Additionally, pictures from 11 persons were not splitted in smaller clusters by YACARE. Choosing $\beta$=0.0 leads of course to a significant amount of data being kept in noise, 46\% here. With $\beta$=$\infty$, one will prefer to have large clusters and no data in the noise. In this case, the amount of pure clusters (with data from a single person) reduced to 17, and two actual clusters were uncut.\\

\begin{table}
\centering
\caption{\textbf{Results for the Olivetti face dataset with different amount of expansion of data.}}
\label{tab:SI-OlivettiData}
\begin{tabular}{ | c || c | c | c | }
\hline
 $\beta$ & Data in the noise (\%) & Pure identified clusters & Uncut actual clusters \\ 
 \hline
 0.0      & 46                  & 35                       & 11 \\  
 1.0      & 33                  & 31                       & 6 \\   
 2.0      & 25                  & 26                       & 6  \\  
 5.0      & 16                  & 23                       & 3  \\  
 10.0     & 15                  & 22                       & 2  \\  
 $\infty$ &  0                  & 17                       & 2  \\  
 \hline
\end{tabular}
\end{table}

We present in Figure~\ref{fig:FigureSI-Olivetti-OtherMethods} the clustering of the Olivetti face dataset data with HDBSCAN and OPTICS with the same minimal cluster size as YACARE i.e. 0.51\% of the data (as expected, when the minimal cluster size was set at 2.0\%, the clustering failed with HDBSCAN and OPTICS since two and nine clusters were found respectively). Results are comparable as what was found for the clustering of kinetics data: HDBSCAN found clusters which were mainly pure and with an acceptable amount of noise (18\%), but found too much clusters (100) with no possibility to easily identify which one should be merged. We note that an advantage of finding too much clusters is that all faces were identified and found in at least one cluster. OPTICS, with default parameters, found a lower amount of clusters (26) which were mainly pure, but with too much noise (61\%). Moreover, 14 persons were not identified, i.e. all their pictures were found in the noise. Thus, for this example, neither HDBSCAN nor OPTICS provided useful results. When we compare the clustering made by YACARE (with $\beta$=0.0) with the one made by density peaks (with data coming from \cite{2014:Rodriguez:Science}) on this dataset, we find comparable results but for different reasons:
\begin{itemize}
    \item Density peaks found 30 clusters, all of them being pure with no mixing of faces in a cluster. YACARE found more clusters (47), 35 of them being pure.
    \item Density peaks identified 26 faces (i.e. 14 faces were found entirely in the noise), whereas YACARE identified up to 38 faces out of 40.
    \item 58\% of data were left in the noise by density peaks, whereas YACARE left aside only 46\% of data with $\beta$=0.0 (and this amount can easily be decreased by changing $\beta$).
    \item Density peaks didn't split a set of photos into two different clusters for 21 faces, whereas this happened for only 11 faces for YACARE. This is a direct consequence of finding more clusters and is not necessarily a problem since YACARE can merge clusters afterwards.
\end{itemize}

\subsubsection*{Clustering of the MNIST database}
The testing set of the MNIST database (10,000 images) was downloaded in .csv format. In this database, each image is in grayscale and is made of 28*28 pixels. Each image was thus described as a vector in dimension 784 where each dimension describes a given pixel and contains an integer between 0 (pure black) and 255 (pure white) (see Figure~\ref{fig:FigureSI-MNIST}-(\textbf{A})). The distance between two images was here computed in a crude way, i.e. as an euclidean distance in dimension 784. With a stencil size of 1.0\%, the plot of $\Delta_d$ appeared highly noisy, we thus increased the stencil size to 2.0\% (see Figures~\ref{fig:FigureSI-MNIST}-(\textbf{B}) and (\textbf{C})).\\

Since we were aware of the difficulty of the problem, we used the minimal cluster size proposed by the algorithm, which is 0.03\% here, in order to capture as much clusters as possible. With the automatically found cut-off, 64 clusters were identified. We then merged the clusters with default parameter ($\alpha$=1.0) starting with the cluster having the fewest number of neighbours, which provided 29 clusters. We expanded clusters with default parameter ($\beta$=1.0), and only 6\% of the data were kept in noise. The confusion matrix is displayed in Figure~\ref{fig:FigureSI-MNIST}-(\textbf{D}). We found that four clusters (out of the 29 found) were pure i.e. contains images from the same digit: clusters \#7, 8 and 9 (which represent digit 1) and cluster \#27 (which represent digit 8 but only contains two elements). Overall 13 clusters describe a given digit by more than 90\%, and 18 clusters describe a given digit by more than 60\%. We must note that we also found that five clusters (\#10, 17, 24, 25 and 28) mixed images from all the ten digits.\\

When the MNIST dataset is clustered with HDBSCAN and OPTICS with a minimal cluster size of 2.0\%, the clustering failed since both methods found a single cluster. When the same minimal cluster size as with YACARE is used (i.e. 0.03\% of the data), both methods also failed: HDBSCAN identified up to 300 clusters with 70\% of the data in the noise, and OPTICS identified 58 clusters with 97\% in the noise (see Figure~\ref{fig:FigureSI-MNIST-OtherMethods}). As explained above, this could be decreased but at the expense of a manual and case-dependent intervention. When this dataset was clustered with the density peaks algorithm, we couldn't identify a cut-off distance $d_c$ that could properly separate the data to identify clusters: in all the decision graphs that were obtained, no clear separation of putative cluster centers was found, making this method unusable (to define the cut-off distance, we tried 0.0008, 0.004, 0.02, 0.1, 0.5 and 2.5 as percentages of the range of distances). Thus, for the difficult case of the MNIST dataset, only YACARE was able to cluster the data.

\subsubsection*{Clustering of experimental conditions from bibliography}
From the NiCOlit database (\cite{2022:Schleinitz:J.Am.Chem.Soc., 2025:NiCOlit:webpage}), we selected six features describing the experimental conditions: the reaction time (in hours), the temperature, and the molar equivalents of the two reactants, the catalyst and the base that were present in the reaction medium. We kept only the reactions for which these six features were all available and well-defined, and excluded cases where values were expressed as ranges. We ended up with 1,620 reactions coming from 37 articles. After standardization of the data, the distances between two reactions were obtained as an euclidean distance in dimension 6 between two vectors.\\

To capture the highest possible number of clusters, we used the minimal possible value of the cluster size (0.13\% here): 61 clusters were found. With $\alpha$=1.0, only two clusters were proposed to be merged and since it wouldn't significantly change the number of clusters we didn't use this option. The confusion matrix with a maximal expansion of clusters ($\beta$=$\infty$) is displayed Figure~\ref{fig:FigureSI-NiCOlit}. We were delighted to observe that the clustering of the reactions based solely on the experimental conditions was highly efficient. Several interesting findings should be highlighted:
\begin{itemize}
    \item The article \#4 was found to be separated in four clusters (\#7 to 10). This article is coming from a collaboration between several groups, coming from four different cities (Taipei and Hsinchu in Taiwan, and Nanjing and Guangzhou in China) \cite{2018:Wang:ACSCatal.}, and we thus hypothesize that each cluster found by YACARE correspond to reactions performed by a different group.
    \item Reactions from article \#7 are divided in several clusters by YACARE, mainly \#12 and 17 (respectively 21 and 49 reactions). In addition to reactions from article \#7, cluster \#12 from YACARE also contains reactions from article \#5, 11 and 32. Thus, there seems to exist a confusion made by YACARE between articles \#5, 7, 11 and 32. However, articles \#7 and 11 are coming from the same laboratory \cite{2012:Muto:J.Am.Chem.Soc.a, 2014:Takise:AngewandteChemieInternationalEditiona}, which explains the confusion.\\
    \item Cluster \#55 from YACARE contains data from articles \#8, 33 and 34, and we found that authors from article \#33 \cite{2015:Nakamura:Org.Lett.} are all coauthors of article \#34 \cite{2017:Schwarzer:J.Am.Chem.Soc.}, which again explains the confusion.\\
\end{itemize}

We clustered the NiCOlit database with other methods. As previously, HDBSCAN found a high number of clusters (170 here) which is too high and non-realistic (see Figure~\ref{fig:FigureSI-NiCOlit-OtherMethods}-(\textbf{A})). OPTICS found a more reasonable number of clusters, though still quite high (90), with an acceptable amount of noise (see Figure~\ref{fig:FigureSI-NiCOlit-OtherMethods}-(\textbf{B})). For both methods, most of the clusters were pure and actual clusters were splitted in smaller ones. For density peaks, choosing a value for $\rho_{min}$ and $\delta_{min}$ from the decision graph was not trivial since no clear separation is found between points (see Figures~\ref{fig:FigureSI-NiCOlit-OtherMethods}-(\textbf{C}-\textbf{D})), whatever the choosen value for the cutoff $d_c$. Using $\rho_{min}$=5.0 and $\delta_{min}$=1.0 lead to the right amount of clusters (37), however only 17 were found to be pure.

\newpage
\subsection*{User Manual}
\subsubsection*{Usage of YACARE with a \textsc{Jupyter} notebook}
\noindent YACARE will be installable with ``pip install yacare\_cluster'' once the manuscript is accepted for publication. Documentation is included in the provided \textsc{Jupyter} notebook, and more details are provided here. An ``autopilot'' function is available, and using automatic parameters it can provide a first quick clustering (see below). To start YACARE, one must first load the modules, create a class to store the variables, define a project name, define the file where the data are stored, and choose to save the images or not.
\begin{verbatim}
import numpy as np
from yacare_cluster import yacare
variables = yacare.Variables()
variables.project_name = "Project"
variables.file_name = "File.dat"
variables.save_images = False
\end{verbatim}

\noindent To load the data, the easiest is to use the provided function \verb"yacare.load_data(variables)". If the data have the same amount of lines than columns and if the diagonal is made with 0s, then we assume the user wants to load a distance matrix. Otherwise, we assume the user is loading a set of features, and we will compute the distance matrix with a euclidean norm. By default, the delimeter between data is ``,'', lines with comments start with `@' or `\#', we load 32-bit floats, and all columns are loaded. In other words, by default we have: \verb|delimiter=",", comments=[`@', `#'], dtype=| \verb"np.float32, usecols=None". This can be changed by the user, for example to skip the first column of the file or to load 64-bit floats (we compared 32-bits and 64-bits and obtained the same results, but some dataset may require more precision). If for some reasons this doesn't work, one can directly load a distance matrix if it was computed before:
\begin{verbatim}
variables.distance_matrix = np.loadtxt(variables.file_name, delimiter=",")
\end{verbatim}
To compute the distances if a set of features was provided, we use the linalg.norm function of \textsc{NumPy}:
\begin{verbatim}
raw_data = np.loadtxt(variables.file_name, delimiter=",", dtype=np.float32)
diffs = raw_data[:, np.newaxis, :] - all_data[np.newaxis, :, :]
variables.distance_matrix = np.linalg.norm(diffs, axis=2)
\end{verbatim}

\noindent If one doesn't have enough RAM to directly compute the matrix, the following lines will work but will take longer to load:
\begin{verbatim}
raw_data = np.loadtxt(variables.file_name, delimiter=",")
number_data = raw_data.shape[0]
variables.distance_matrix = np.zeros((number_data, number_data))
for i in range(number_data):
    diffs = raw_data[i] - raw_data
    variables.distance_matrix[i] = np.linalg.norm(diffs, axis=1)
\end{verbatim}
and the following lines will take even longer but should fit any modern hardware:
\begin{verbatim}
for i in range(raw_data.shape[0]):
    for j in range(i, raw_data.shape[0]):
        dist = np.linalg.norm(data[i] - data[j])
        variables.distance_matrix[i, j] = dist
        variables.distance_matrix[j, i] = dist
\end{verbatim}
Once the distance matrix is computed, it can be saved as a binary file to load it more rapidly:
\begin{verbatim}
np.save(variables.project_name + "_Matrix.npy", variables.distance_matrix)
\end{verbatim}

\noindent The main variables that need to be defined during the usage of YACARE are the following ones:
\begin{verbatim}
percentage_moving_square = 1.0  # Stencil size
minimal_size_cluster = 2.0      # Minimal cluster size
threshold_variable = 1.0        # Defined as alpha above
amount_of_noise = 1.0           # Defined as beta above
\end{verbatim}

\noindent YACARE can then be used simply with the following functions:
\begin{verbatim}
yacare.perform_first_reordering(variables, percentage_moving_square)
yacare.choose_if_we_reorder_again(variables)
yacare.find_optimal_cutoff(variables, minimal_size_cluster)
yacare.change_proposed_cutoff(variables)
yacare.find_final_clusters(variables)
yacare.compare_clusters(variables) 
yacare.propose_list_for_concatenating_clusters(variables, threshold_variable)
yacare.concatenate_clusters(variables)
yacare.expand_clusters(variables, amount_of_noise)
yacare.compare_final_clusters(variables)
yacare.write_indices(variables)
\end{verbatim}

\noindent Only \verb"perform_first_reordering", \verb"find_optimal_cutoff" and \verb"find_final_clusters" are mandatory. A few additional variables can be changed from their default values to refine the analysis or the presentation of the results if needed.
\begin{itemize}
    \item If one wants to use the function \verb"choose_if_we_reorder_again" in command-line, the parameter \verb"indices=[...]" can be added, where the list of indices to try for reordering is given (this will skip the question to know which type of reordering to do). This can be useful if one has a big distance matrix and wants to parallelize the search for the best reordering point. For example, to search from 0 to 5000 use \verb"indices=list(range(0,5001))", and to search from 5000 to 10000 use \verb"indices=list(range(5000,10001))".
    \item When searching for the optimal cutoff, one can use \verb"use_all_cutoff=False" in the function \verb"find_optimal_cutoff" if the plot of $\Delta_d$ is noisy and one wants to avoid looking for cut-offs in the lowest 10\% of the values (by default, this parameter is set to \verb"True").
    \item If the plot of $\mathcal{S}$ with respect to cut-off doesn't display a clear peak (as in Figure~\ref{fig:Figure1}-(\textbf{C})), one can use the parameter \verb"function_for_ratio=2" to use another mathematical function in the function \verb"find_optimal_cutoff" (by default, this parameter is set to \verb"1").
    \item The functions \verb"compare_clusters" and \verb"compare_final_clusters" can be feeded with the parameter \verb"display_stddev=True" and/or \verb"display_mean_distances=True" if ones wants to display the standard deviations and/or the mean distance in each zone in the image as is shown in Figure~\ref{fig:FigureSI-Tools}-(\textbf{B}) (by default, they are set to \verb"False").
    \item If one wants to directly specify how to merge clusters with options \verb"1", \verb"2", \verb"3" or \verb"4", the parameter \verb"choice_merging_clusters" can be added to the function \verb"propose_list_for_concatenating" \verb"_clusters" (by default, it is set to \verb"0"). These four options correspond respectively to start merging from the smallest cluster, to start merging from the largest cluster, to start merging from the cluster with the highest number of neighbours, or to start merging from the cluster with the fewest number of neighbours.
    \item If one wants to use $\beta$=$\infty$ when expanding the clusters, the parameter \verb"keep_no_noise=True" must be used in the function \verb"expand_clusters" (in that case, the value of \verb"amount_of_noise" doesn't matter).
    \item If you have outliers, they may pollute the presentation of the distance matrices. Thus, you may want to restrict the range of data in the colorbar. To do so, you can add the parameter \verb"vmax=..." in all functions that display a distance matrix. By default, \verb"vmax" is the maximal value in the distance matrix and you don't need to give it if you are happy with the images.
\end{itemize}

\subsubsection*{Validation of the clustering}
\noindent We have included in the module the option to compute a confusion matrix with YACARE, HDBSCAN, OPTICS and k-means (using the \textsc{Scikit-learn} implementation for the last three algorithms), as shown in the following lines. There is no reason why the cluster ``2" from YACARE should be the same as cluster ``2" from the true value; clusters found by YACARE can thus be reordered to make a prettier image with large clusters closer to the diagonal. This is done automatically with the parameter \verb"auto_reorder_columns=True" which will only reorder the columns and will provide in the output the performed reordering in the form of a dictionary. This parameter can be used for the four clustering algorithms. The dictionary can be used as a parameter \verb"transformation = {...}" to reorder the columns in a controlled way. Note that the \textit{k}-means algorithm should only be used when the user provides a set of features for each point.
\begin{verbatim}
labels_true = np.loadtxt("Labels.csv", comments=["@", "#"], usecols=0)
yacare.plot_confusion_matrix(variables, labels_true, auto_reorder_columns=True)
yacare.plot_confusion_matrix_HDBSCAN(variables, labels_true)
yacare.plot_confusion_matrix_OPTICS(variables, labels_true)
yacare.plot_confusion_matrix_kmeans(variables, labels_true)
\end{verbatim}
After running these functions, the predicted labels can be accessed with \verb"variables.labels_yacare", \verb"variables.labels_hdbscan", \verb"variables.labels_optics", \verb"variables.labels_kmeans" and \verb"variables.labels_density_peaks".

\subsubsection*{Usage of YACARE in command line}
\noindent To use YACARE in command line through a script, one can adapt the following lines.
\begin{verbatim}
import numpy as np
from yacare_cluster import yacare
variables = yacare.Variables()
variables.project_name = "Project"
variables.file_name = "File.csv"
variables.save_images = True
yacare.load_data(variables)
yacare.perform_first_reordering(variables, percentage_moving_square = 1.0)
yacare.choose_if_we_reorder_again(variables, indices=list(range(0,
                                                  variables.num_elements)))
yacare.find_optimal_cutoff(variables, minimal_size_cluster = 2.0)
yacare.find_final_clusters(variables)
yacare.propose_list_for_concatenating_clusters(variables, threshold_variable
                                        = 1.0, choice_merging_clusters = 3)
yacare.concatenate_clusters(variables)
yacare.expand_clusters(variables, amount_of_noise = 1.0)
yacare.compare_final_clusters(variables)
yacare.write_indices(variables)
labels_true = np.loadtxt("Labels.csv", comments=["@", "#"], usecols=0)
yacare.plot_confusion_matrix(variables, labels_true, auto_reorder_columns=True)
\end{verbatim}

\subsubsection*{Usage of YACARE in autopilot}
\noindent We have included a function that runs everything automatically. It can be called with:
\begin{verbatim}
from yacare_cluster import yacare
variables = yacare.Variables()
yacare.autopilot(variables, project_name = "Project", file_name = "File.csv")
\end{verbatim}
This function has several additional parameters that can be used. They have been already described, and we here only provide their default values:
\begin{verbatim}
save_images = False
percentage_moving_square = 1.0
indices=[]   # Meaning that all elements will be tried for reordering
minimal_size_cluster = 2.0
function_for_ratio = 1
threshold_variable = 1.0
choice_merging_clusters = 3
amount_of_noise = 1.0
keep_no_noise = False
\end{verbatim}

\subsubsection*{Study of biomolecules - Distance matrix}
\noindent When one wants to study biomolecules and use the RMSD as the distance between two elements in the data, the following command can be used with Gromacs to create the distance matrix:
\begin{verbatim}
echo "Protein Protein" | gmx rms -s MD.tpr -n Index.ndx -f MD.xtc -f2 MD.xtc 
-o MD_RMSDMatrix.xvg -m MD_RMSDMatrix.xpm -bin MD_RMSDMatrix.dat
\end{verbatim}
The file can then be read by \textsc{Numpy} with:
\begin{verbatim}
file_size = os.path.getsize("MD_RMSDMatrix.dat")
element_size = np.dtype(np.float32).itemsize
total_elements = file_size // element_size
matrix_size = int(total_elements**0.5)
data = np.fromfile(File, dtype=np.float32)
variables.distance_matrix = data.reshape((matrix_size, matrix_size))
\end{verbatim}
Another option is to use MDAnalysis:
\begin{verbatim}
import numpy as np
import MDAnalysis as mda
from MDAnalysis.analysis import rms
u = mda.Universe("MD.pdb", "MD.xtc")         #Load files
selection = u.select_atoms("protein")        #Select atoms
n_frames = len(u.trajectory[::10])           #Number of frames, 1 every 10
rmsd_matrix = np.zeros((n_frames, n_frames)) #Initialize matrix
#Compute the distance matrix
for i, ts1 in enumerate(u.trajectory[::10]):
    ref_positions = selection.positions.copy()
    for j, ts2 in enumerate(u.trajectory[::10]):
        if j >= i:
            rmsd = rms.rmsd(selection.positions, ref_positions, center=True,
                   superposition=True)
            rmsd_matrix[i, j] = rmsd
            rmsd_matrix[j, i] = rmsd
np.save("rmsd_matrix.npy", rmsd_matrix)  #Save file
\end{verbatim}

\subsubsection*{Study of biomolecules - Reordering a trajectory}
\noindent If one wants to reorder a Gromacs trajectory according to the order found by YACARE, the following command can be used:
\begin{verbatim}
gmx extract-cluster -clusters Project_Clustering_Clusters.ndx -f MD.xtc 
\end{verbatim}
This will output a series of file (\textit{trajout\_Cluster*.xtc} for the clusters, and \textit{trajout\_Noise.xtc} for the noise), which can then be concatenated with:
\begin{verbatim}
gmx trjcat -f trajout_Cluster0.xtc trajout_Cluster1.xtc ... trajout_Noise.xtc
                                   -o MD_Reordered.xtc -sort no -cat yes
\end{verbatim}

\newpage


\begin{figure}
	\centering
	\includegraphics[width=1.0\textwidth]{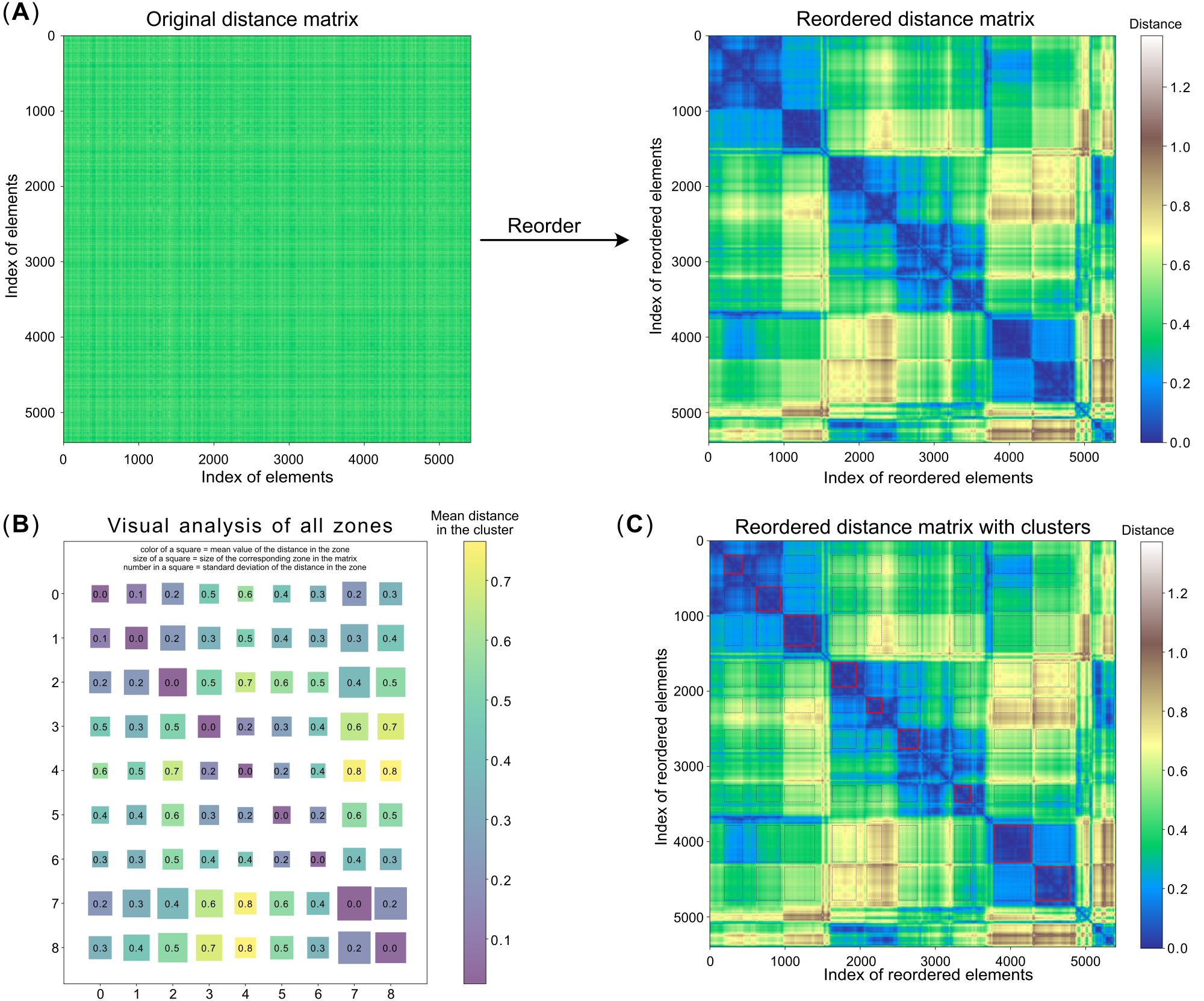}
	\caption{\textbf{Reordering a matrix, and proposed tools to compare clusters.} (\textbf{A}) Original and reordered distance matrices from the toy dataset. (\textbf{B}) The size, mean distance and standard deviation of the distance in all clusters and off-diagonal zones are displayed. The user can choose to display in each square the mean distance, the standard deviation, both, or none. (\textbf{C}) All clusters are identified as red squares, and off-diagonal zones as dashed-purple rectangles. The values for the zones on panel (\textbf{B}) match the zones on panel (\textbf{C}).}
	\label{fig:FigureSI-Tools}
\end{figure}

\begin{figure}
	\centering
	\includegraphics[width=1.0\textwidth]{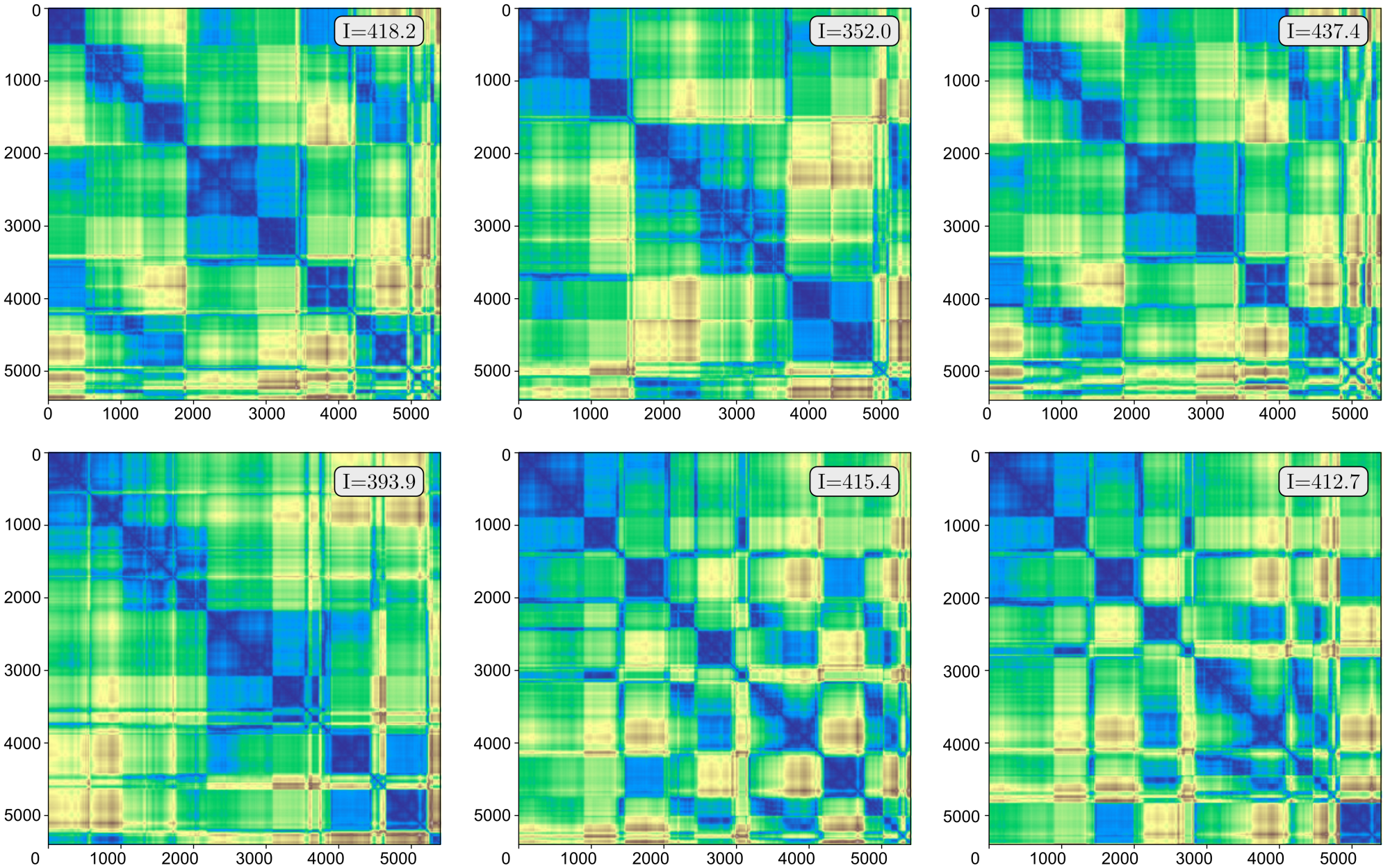}
	\caption{\textbf{Comparing different starting points for reordering the distance matrix.} The reordering of the distance matrix with six starting points is compared, and the integral of $\Delta_d$ is provided. First row: reordering according to the first element, to the element that provides the lowest value for the integral of $\Delta_d$, to the element that provides the highest value for the integral of $\Delta_d$. Second row: reordering according to three random elements.}
	\label{fig:FigureSI-CompareReordering}
\end{figure}

\begin{figure}
	\centering
	\includegraphics[width=1.0\textwidth]{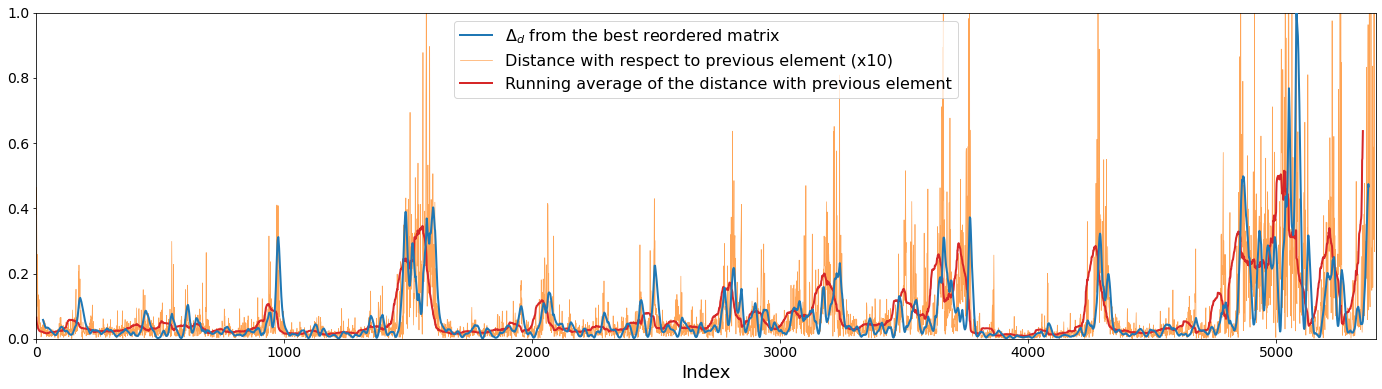}
	\caption{\textbf{Comparing distances.} Super-imposition of $\Delta_d$ (in blue, computed with 1\% of the dataset i.e. 54 elements), of the sequential distances with respect to the first neighbour (i.e. the first off-diagonal of the distance matrix, in orange), and of the running average of the sequential distance with respect to the first neighbour (computed over 50 elements, in red). Data are coming from the toy dataset.}
	\label{fig:FigureSI-CompareDistances}
\end{figure}

\begin{figure}
	\centering
	\includegraphics[width=1.0\textwidth]{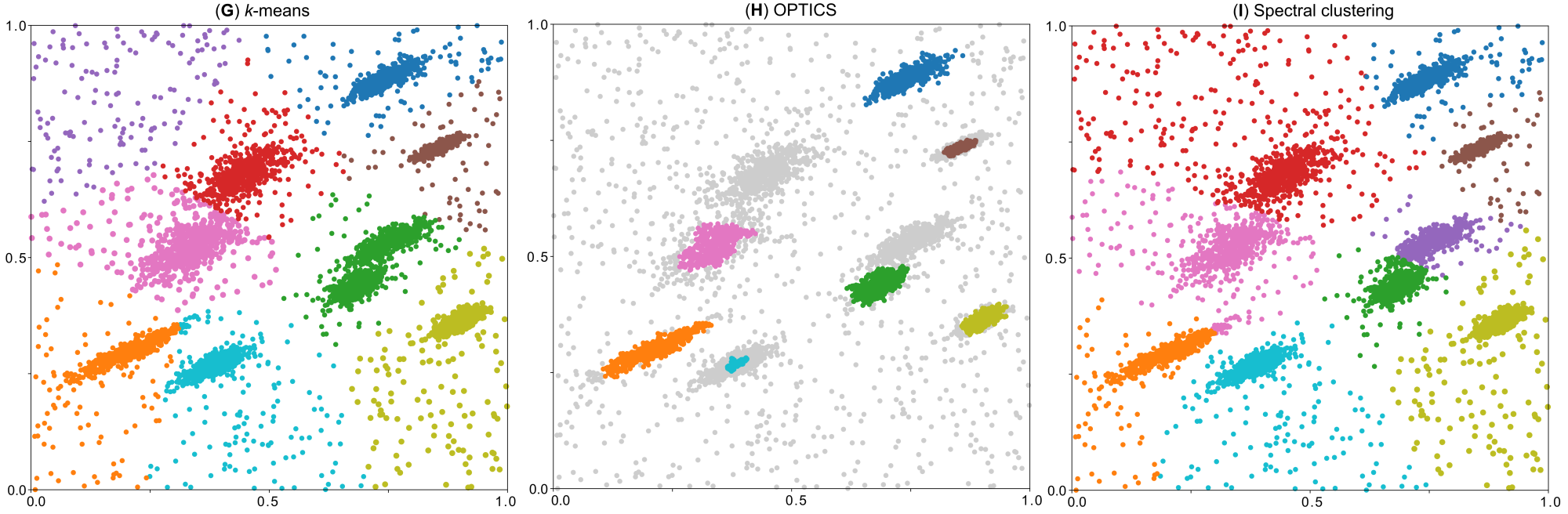}
	\caption{\textbf{Comparison of methods.} (\textbf{G}) Clustering with the \textit{k}-means method, asking for nine clusters, (\textbf{H}) Clustering with the OPTICS methods, asking for a minimal cluster size corresponding to 2.0\% of the data, (\textbf{I}) Clustering with spectral clustering, asking for nine clusters.}
	\label{fig:FigureSI-OtherMethods}
\end{figure}

\begin{figure}
	\centering
	\includegraphics[width=1.0\textwidth]{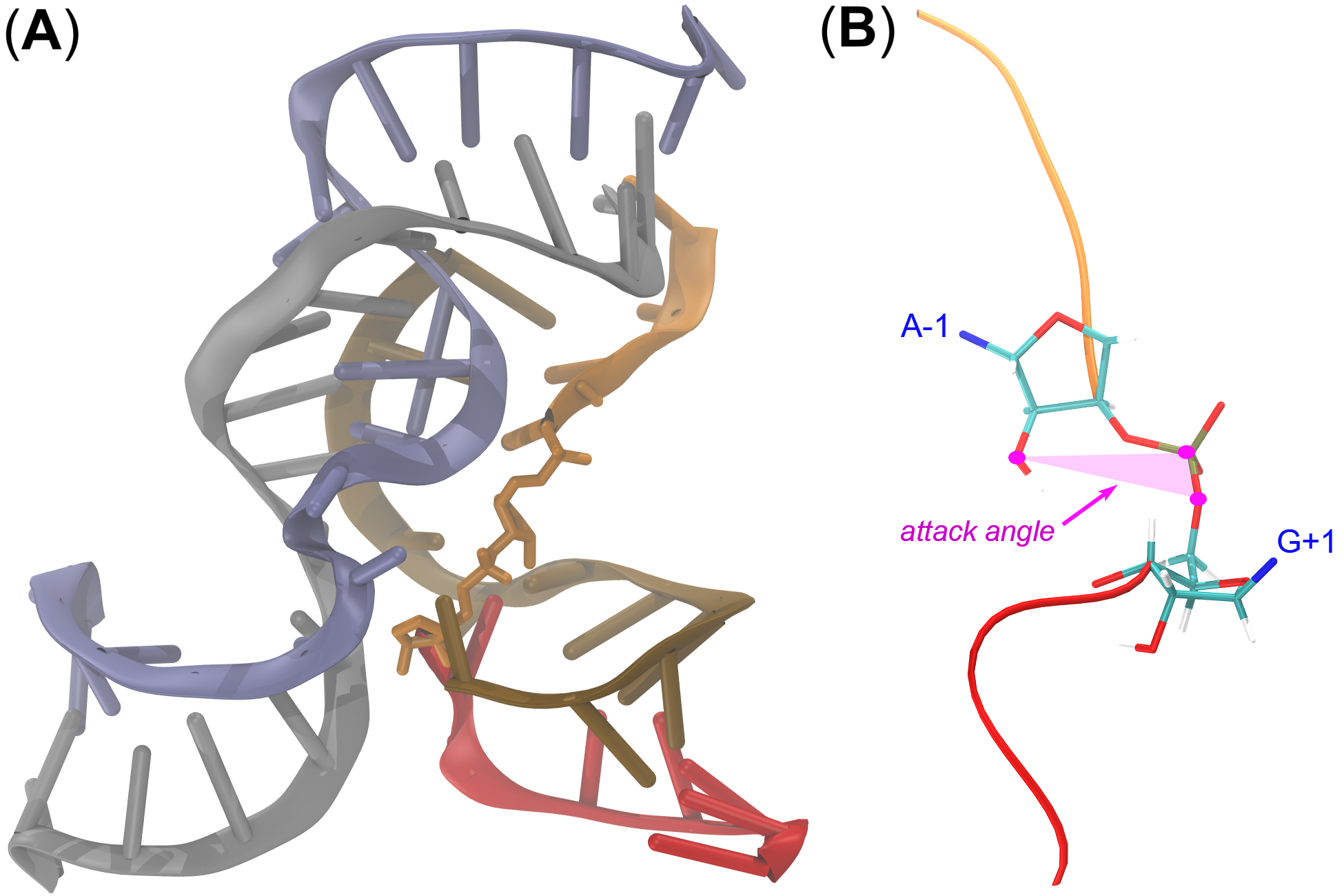}
	\caption{\textbf{Structure of the RNA ribozyme.} (\textbf{A}) Global view of the hairpin ribozyme, with each RNA strand colored differently. The catalytic strand is depicted in orange and red, with the cleavable bond located at the junction between these two colors. The catalytic residues, A-1 (orange) and G+1 (red), are represented in licorice style. (\textbf{B}) Close-up of the cleavable junction between A-1 and G+1. The attack angle, which must be flat for the reaction to proceed, is highlighted in pink.}
	\label{fig:FigureSI-ARN-Structure}
\end{figure}

\begin{figure}
	\centering
	\includegraphics[width=0.95\textwidth]{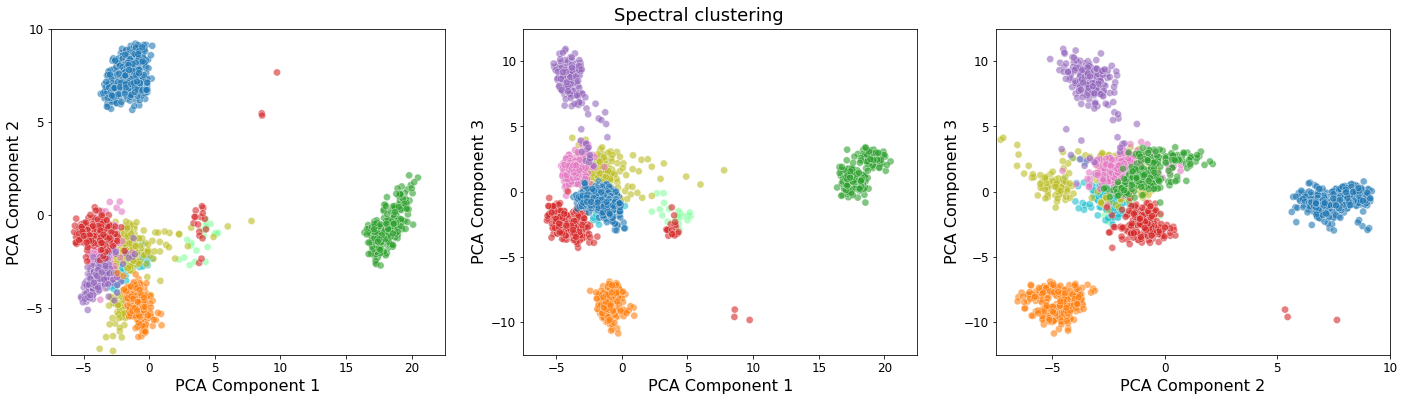}
	\includegraphics[width=0.95\textwidth]{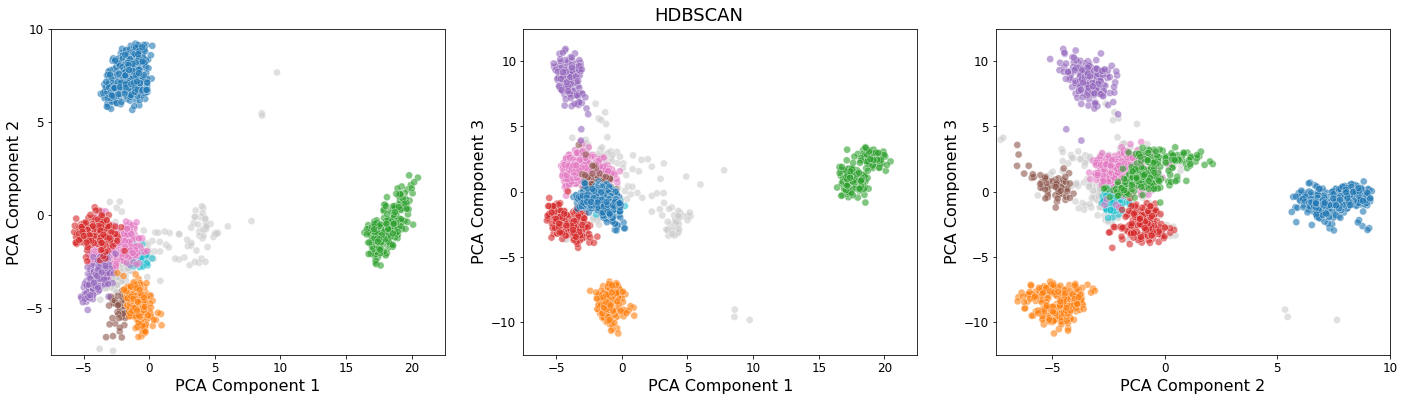}
	\includegraphics[width=0.95\textwidth]{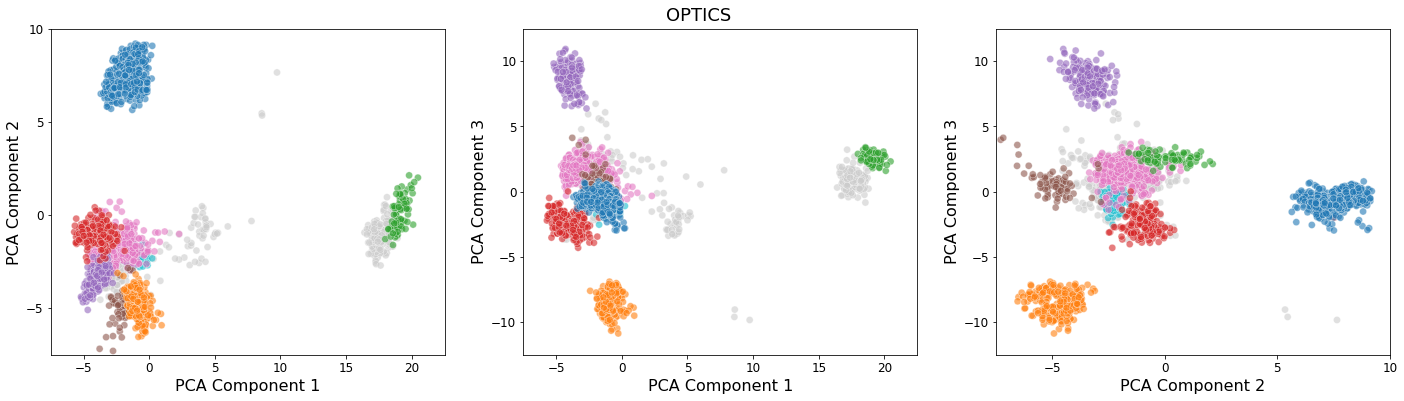}
	\includegraphics[width=0.95\textwidth]{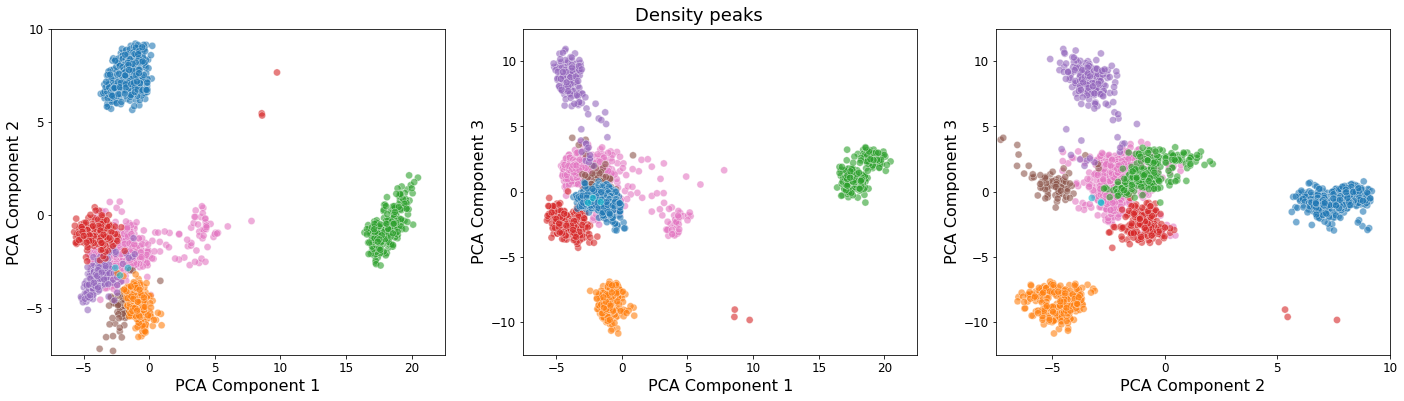}
	\caption{\textbf{Comparison between various clustering methods and PCA.} Data points from the RNA structure are plotted on two PCA components (first \textit{vs} second, first \textit{vs} third, second \textit{vs} third) and are colored according to the found clusters by the method. For spectral clustering, nine clusters were asked; for HDBSCAN and OPTICS, the minimal cluster size was set at 2.0\% of the data; for density peaks, $\rho_{min}$ and $\delta_{min}$ were chosen according to the decision graph.}
	\label{fig:FigureSI-ARN-OtherMethods}
\end{figure}

\begin{figure}
	\centering
	\includegraphics[width=0.9\textwidth]{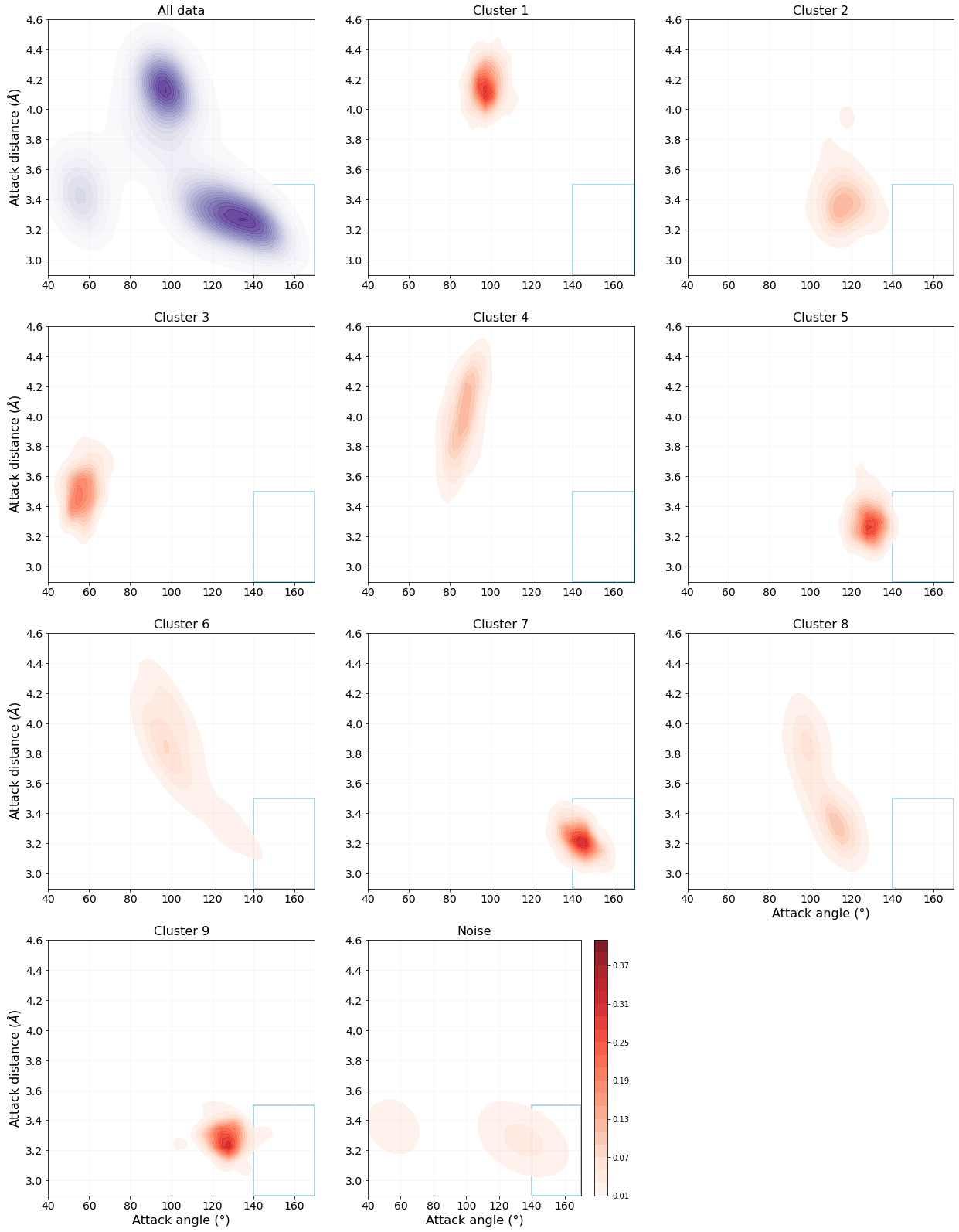}
	\caption{\textbf{Angles and distances for the RNA structures.} Values of the angles and distances (shown as densities in 2D) for the 2500 snapshots of the RNA structure. The full set of data is first presented in purple, and the separation of data in each cluster is then shown in red. The blue rectangle indicates the reactive geometries: only cluster \#7 provides meaningful structures.}
	\label{fig:FigureSI-ARN-AnglesDistances}
\end{figure}

\begin{figure}
	\centering
	\includegraphics[width=1.0\textwidth]{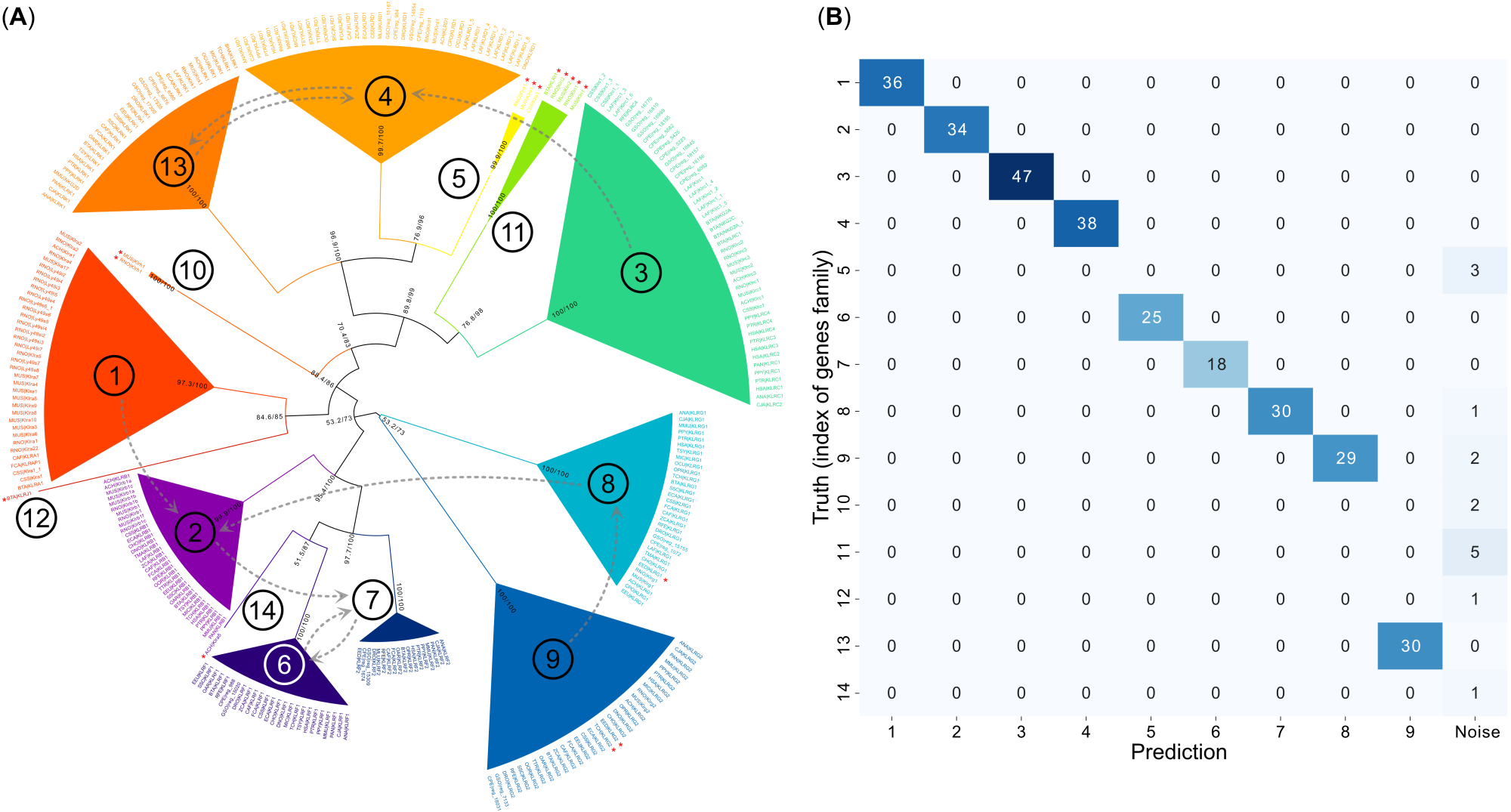}
	\caption{\textbf{Clustering of genes.} (\textbf{A}) Phylogenetic tree for the killer cell lectin-like receptor (KLR) obtained with FigTree. Circled numbers indicates the index of the cluster of genes. Genes labelled with a red star were found in the noise with YACARE (with $\beta$=1.0), and gray arrows indicates the closest cluster from each cluster. Bootstrap values are displayed in each branch. (\textbf{B}) Confusion matrix for the clustering of KLR genes.}
	\label{fig:FigureSI-Genes}
\end{figure}

\begin{figure}
	\centering
	\includegraphics[width=0.32\textwidth]{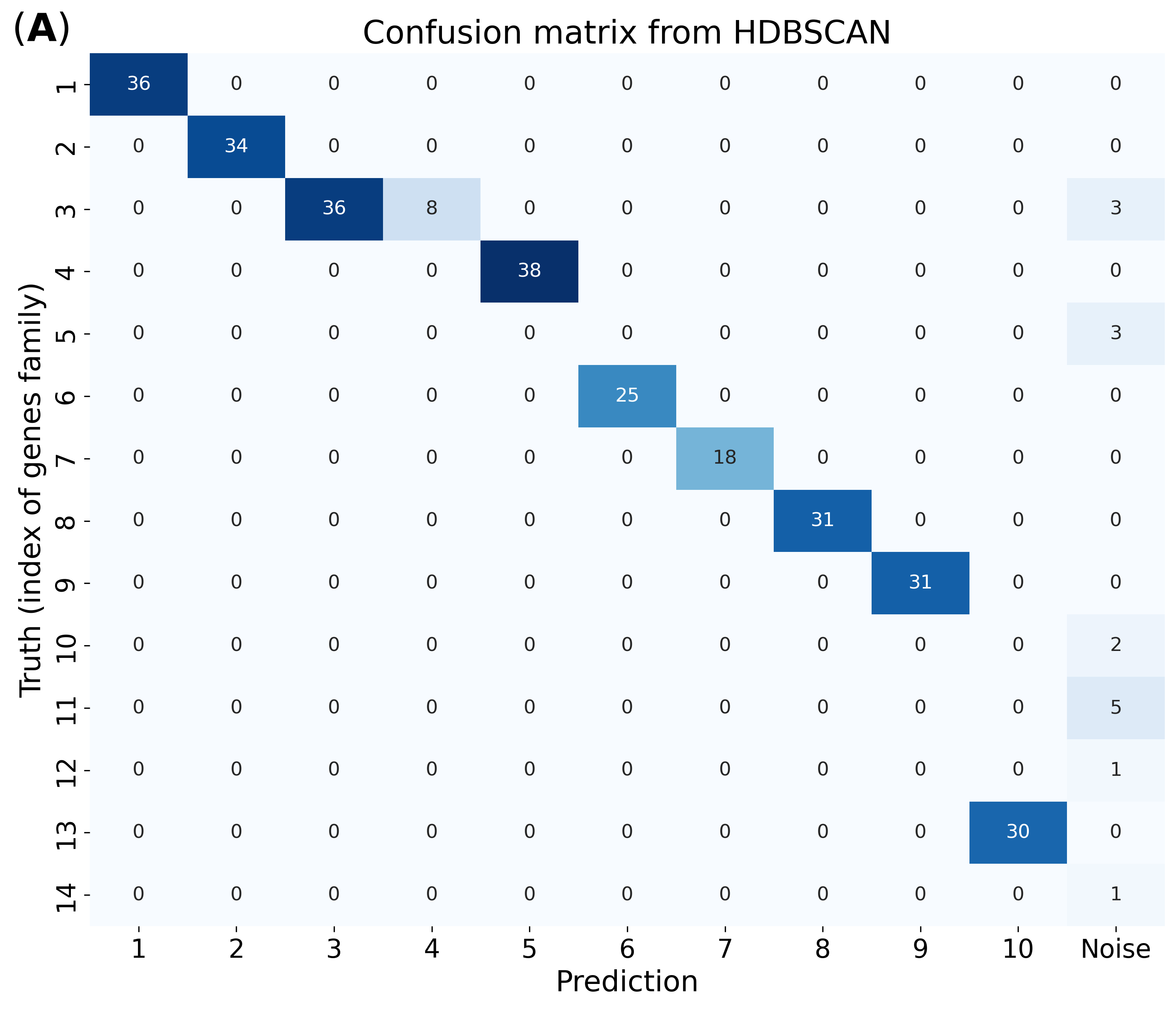}
	\includegraphics[width=0.32\textwidth]{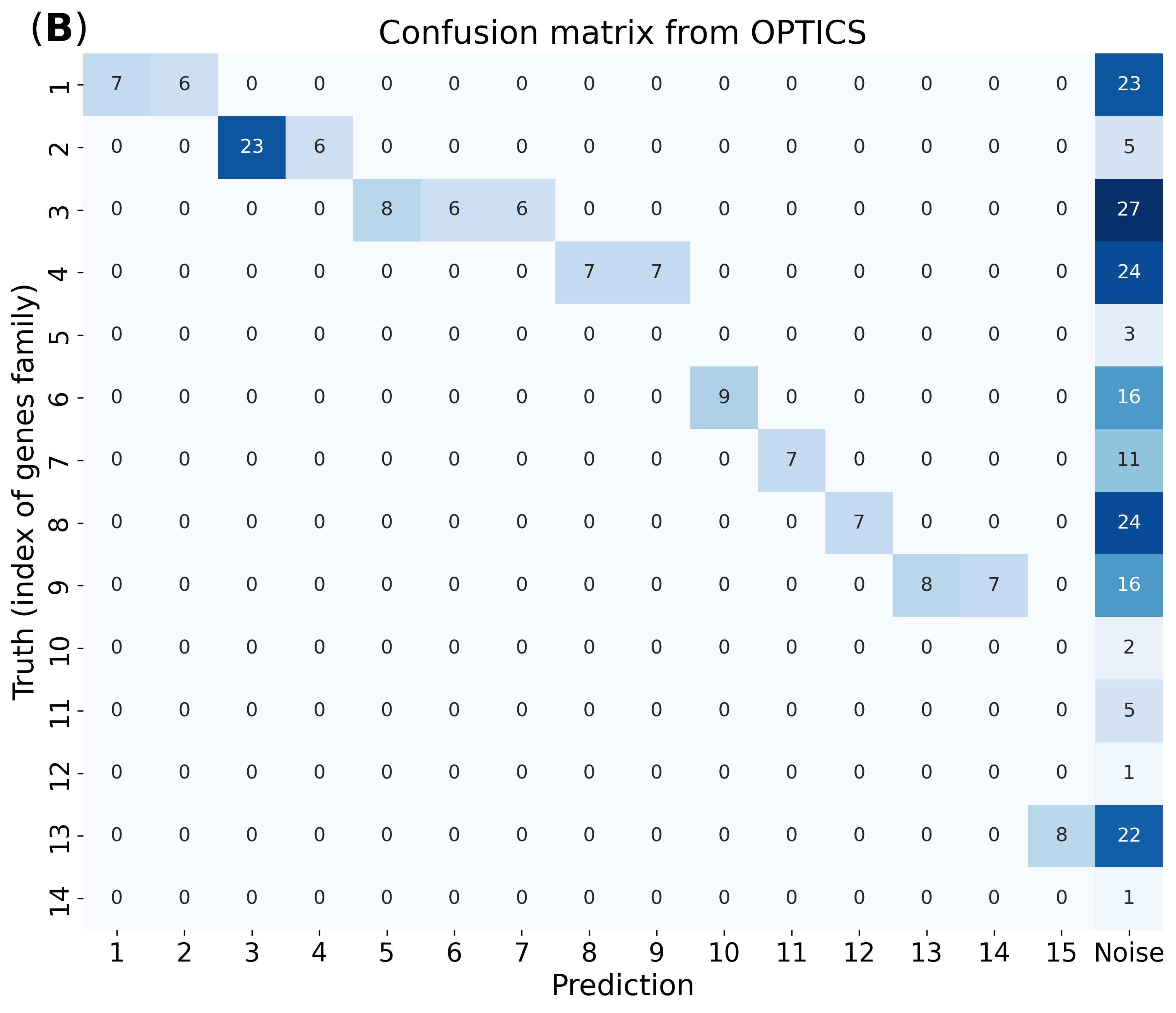}
	\includegraphics[width=0.32\textwidth]{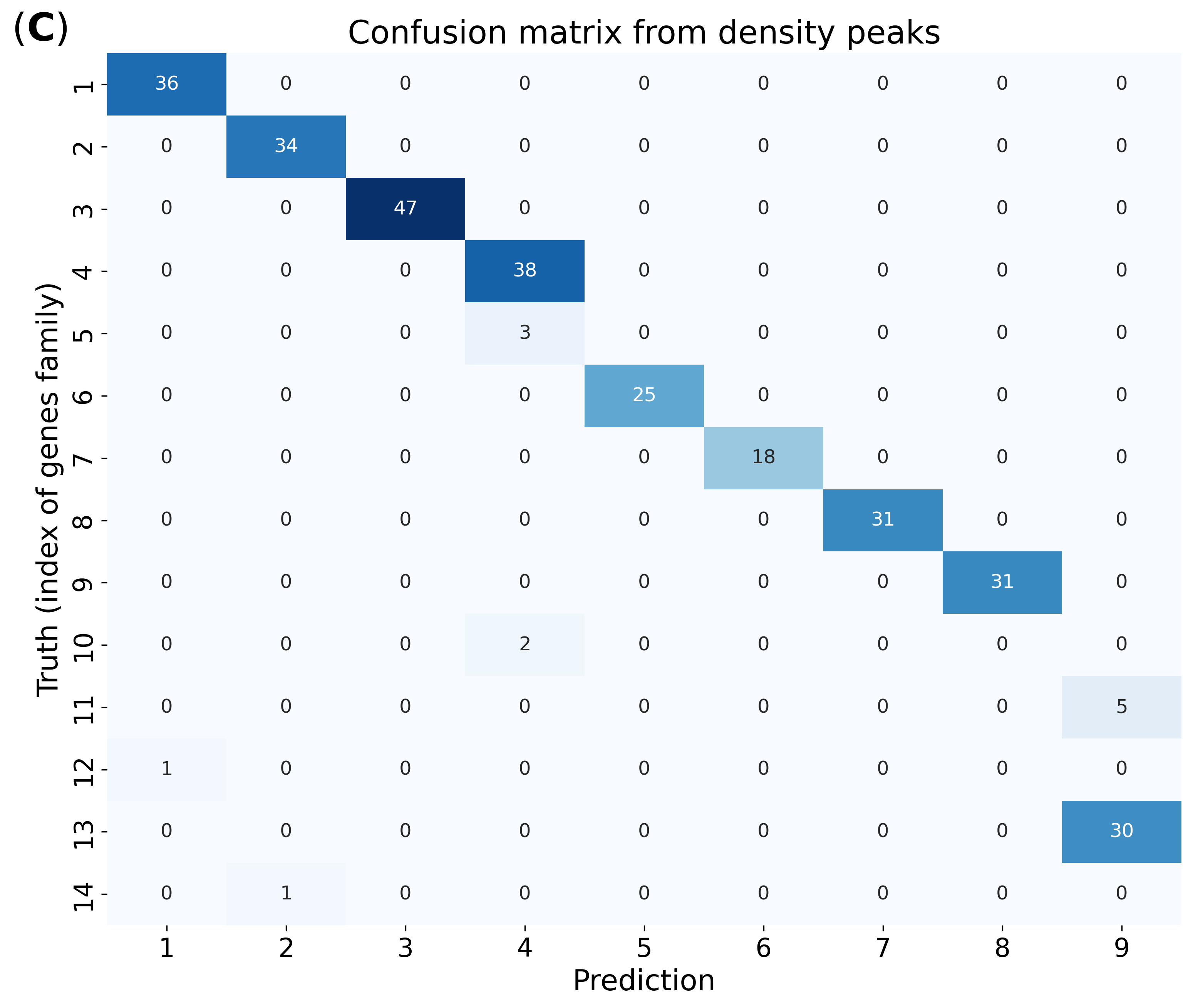}
	\caption{\textbf{Confusion matrices for the clustering of genes.} (\textbf{A}) With HDBSCAN with a minimal cluster size of 2.0\% of the data. (\textbf{B}) With OPTICS with a minimal cluster size of 2.0\% of the data. (\textbf{C}) With density peaks with $\rho_{min}$ and $\delta_{min}$ chosen according to the decision graph.}
	\label{fig:FigureSI-Genes-OtherMethods}
\end{figure}

\begin{figure}
	\centering
	\includegraphics[width=0.75\textwidth]{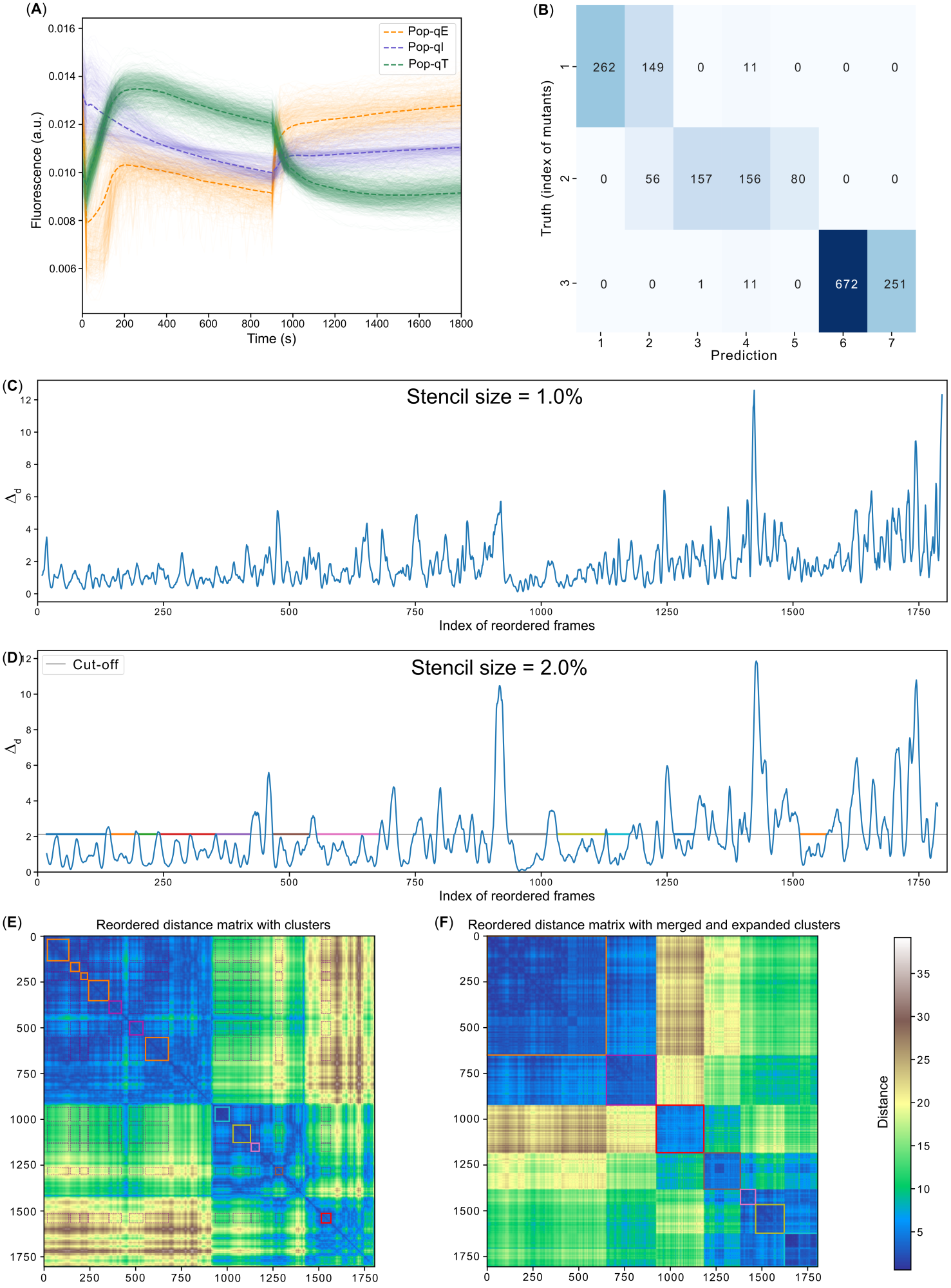}
	\caption{\textbf{Clustering of fluorescence kinetic data.} (\textbf{A}) Kinetics of fluorescence: pale lines are the raw data, and dashed lines are averaged data. (\textbf{B}) Confusion matrix with a full expansion of clusters (i.e. $\beta$=$\infty$). (\textbf{C}) Plot of $\Delta_d$ with a stencil size of 1.0\%. (\textbf{D}) Plot of $\Delta_d$ with a stencil size of 2.0\% and the 12 identified clusters. (\textbf{E}) Automatically found clusters displayed in the best reordered matrix. (\textbf{F}) Final clusters after merging and expansion. Clusters with the same colors in panel (\textbf{E}) were merged to the corresponding clusters in panel (\textbf{F}).}
	\label{fig:FigureSI-Kinetics}
\end{figure}

\begin{figure}
	\centering
	\includegraphics[width=1.0\textwidth]{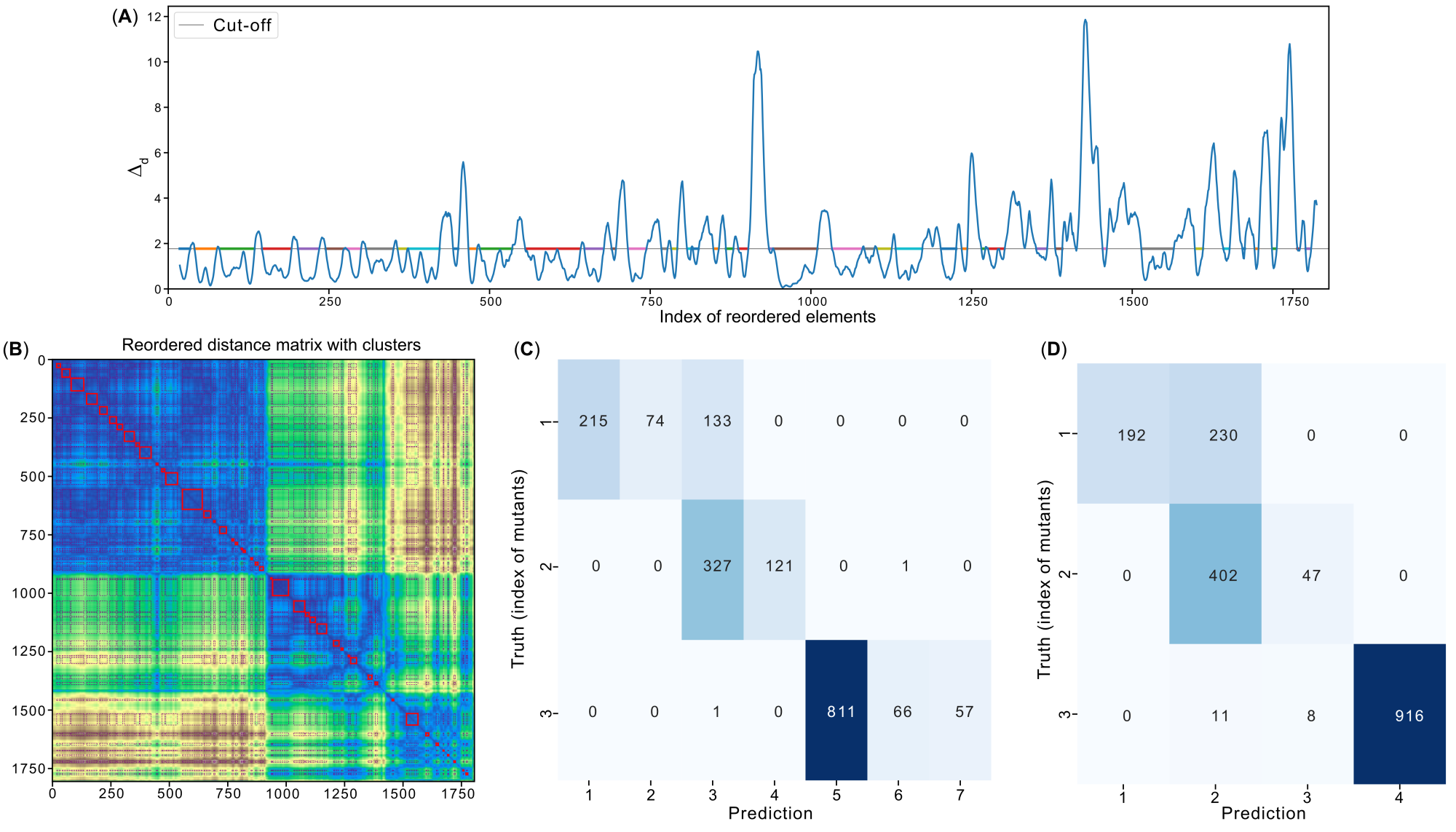}
	\caption{\textbf{Clustering of fluorescence kinetic data with a smaller minimal size.} (\textbf{A}) Plot of $\Delta_d$ with a stencil size of 0.12\% and the 45 identified clusters. (\textbf{B}) Automatically found clusters displayed in the best reordered matrix. (\textbf{C}) Confusion matrix after automatic merging ($\alpha$=2.0) and expansion ($\beta$=$\infty$) of clusters. (\textbf{D}) Confusion matrix after automatic merging ($\alpha$=3.0) and expansion ($\beta$=$\infty$) of clusters.}
	\label{fig:FigureSI-Kinetics-SmallerSize}
\end{figure}	

\begin{figure}
	\centering
	\includegraphics[scale=0.35]{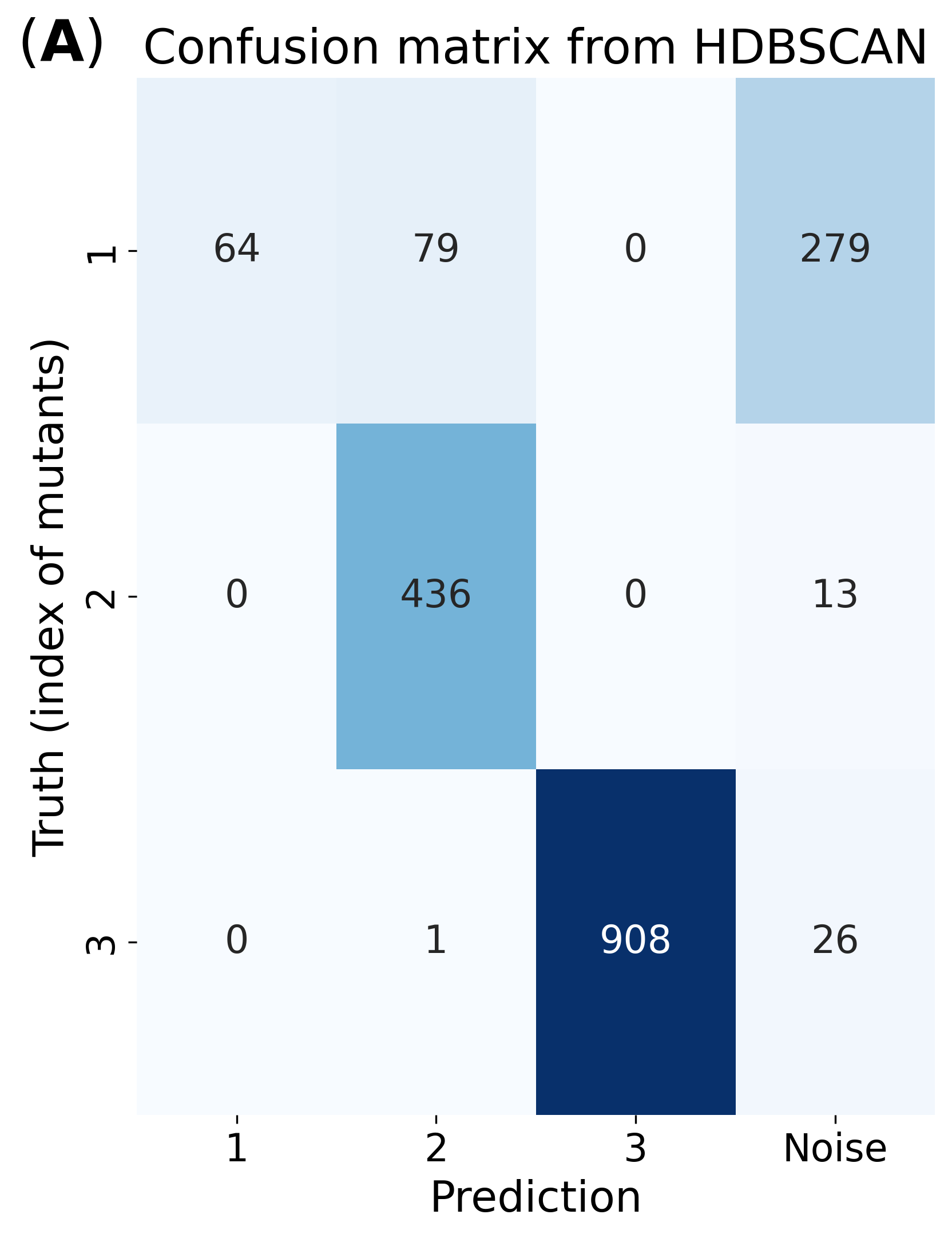}\hspace{1cm}
	\includegraphics[scale=0.35]{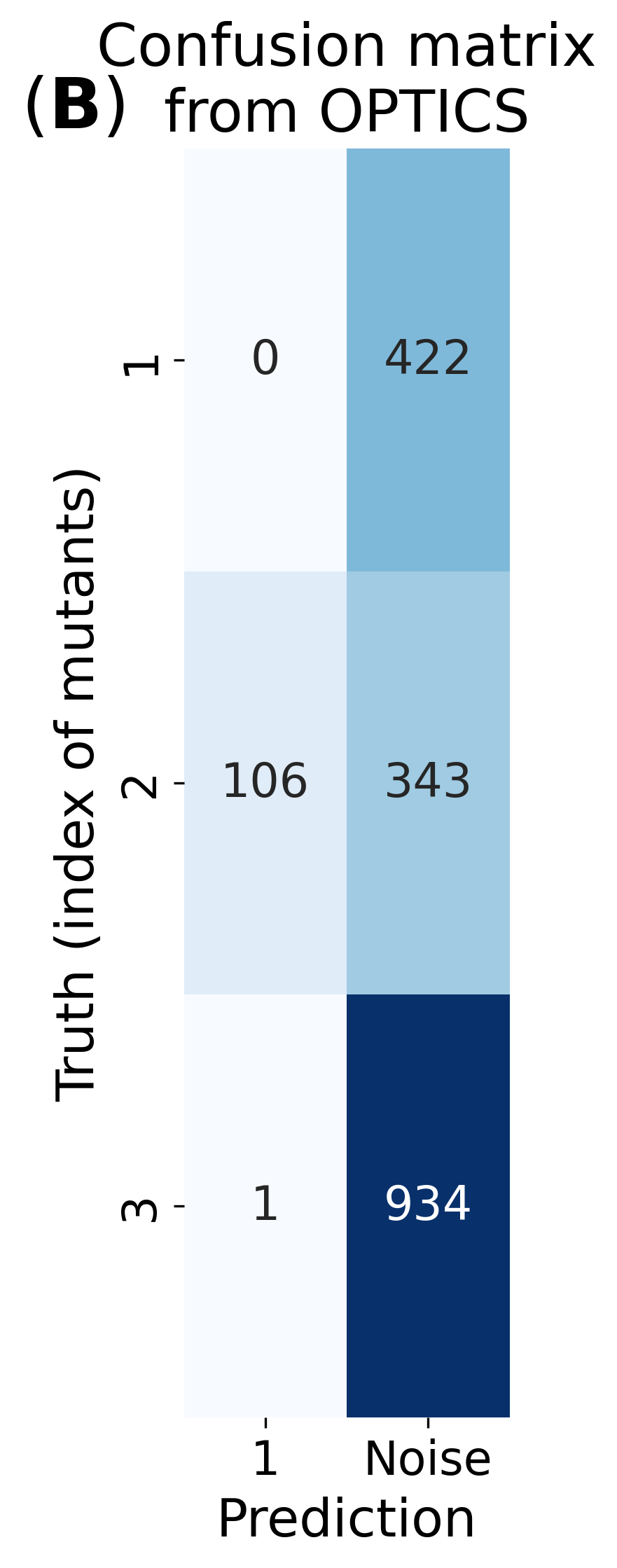}\hspace{1cm}
	\includegraphics[scale=0.35]{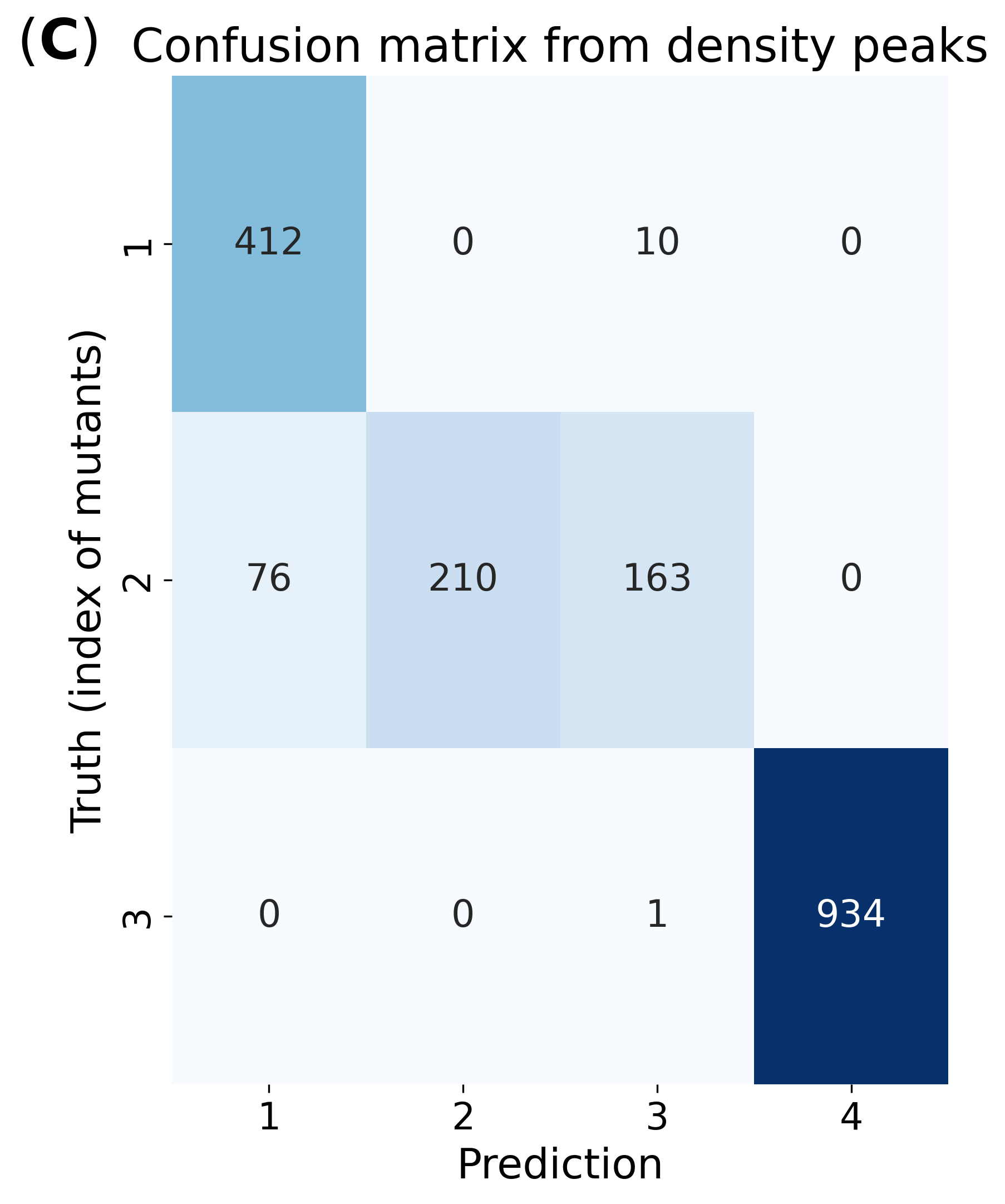} \\
	\vspace{0.5cm}
	\includegraphics[width=1.0\textwidth]{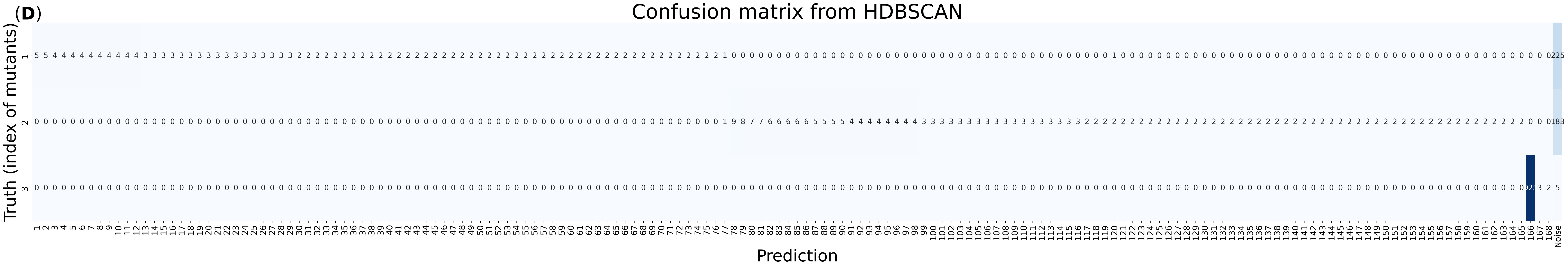} \\
	\vspace{0.5cm}
	\includegraphics[width=0.8\textwidth]{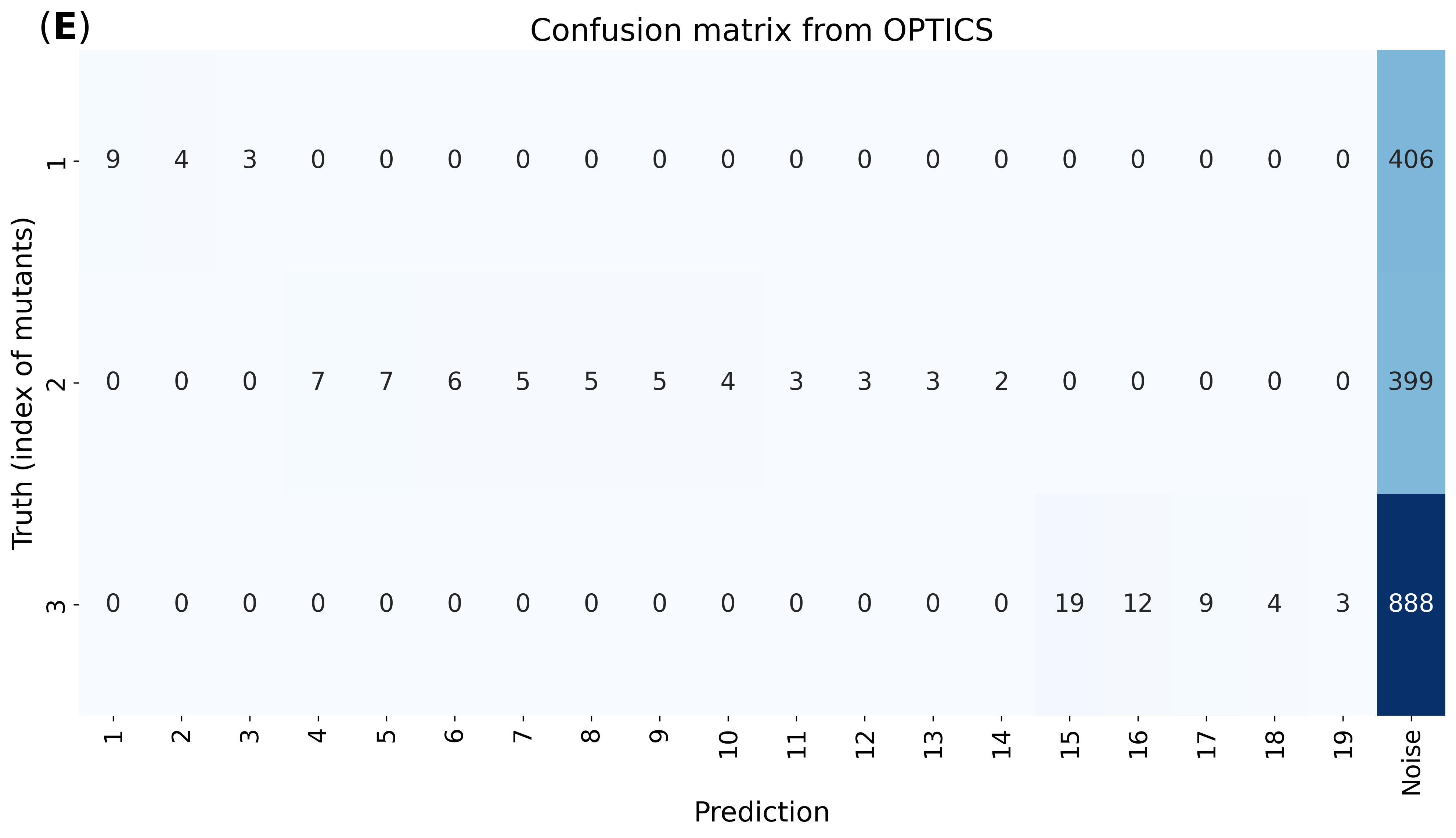}
	\caption{\textbf{Confusion matrices for the clustering of kinetics data.} (\textbf{A}) With HDBSCAN with a minimal cluster size of 2.0\% of the data. (\textbf{B}) With OPTICS with a minimal cluster size of 2.0\% of the data. (\textbf{C}) With density peaks with $\rho_{min}$ and $\delta_{min}$ chosen according to the decision graph. (\textbf{D}) With HDBSCAN with a minimal cluster size of 0.12\% of the data. (\textbf{E}) With OPTICS with a minimal cluster size of 0.12\% of the data.}
	\label{fig:FigureSI-Kinetics-OtherMethods}
\end{figure}

\begin{figure}
	\centering
	\includegraphics[width=0.95\textwidth]{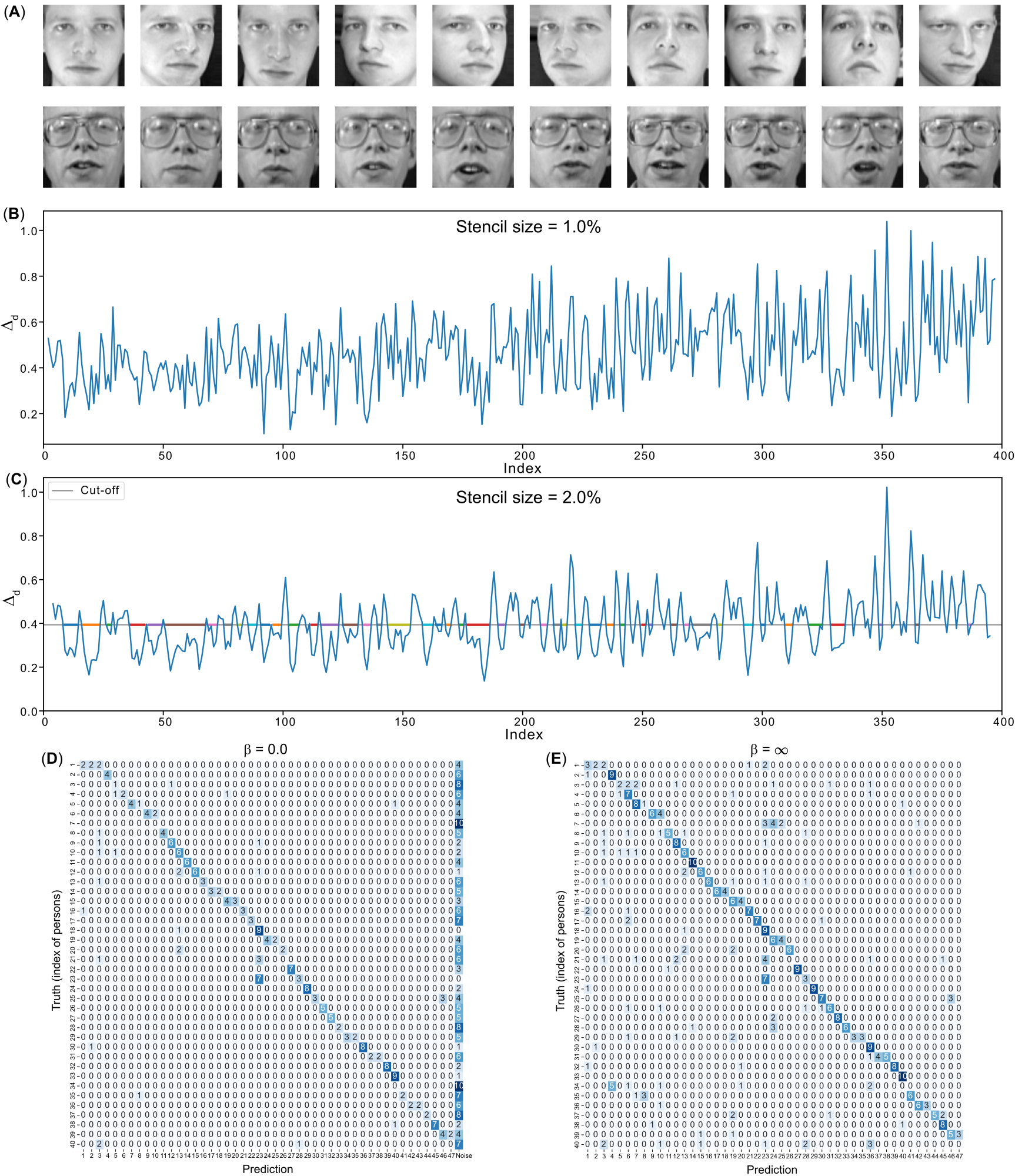}
	\caption{\textbf{Clustering of the Olivetti face dataset.} (\textbf{A}) Twenty first images (ten photos for two persons) from the Olivetti database. (\textbf{B}) Plot of $\Delta_d$ with a stencil size of 1.0\%. (\textbf{C}) Plot of $\Delta_d$ with a stencil size of 2.0\%, and found clusters with the automatically proposed cut-off. (\textbf{D}) Confusion matrix with a minimal expansion of clusters (i.e. $\beta$=0.0). (\textbf{E}) Confusion matrix with a full expansion of clusters (i.e. $\beta$=$\infty$).}
	\label{fig:FigureSI-Olivetti}
\end{figure}

\begin{figure}
	\centering
	\includegraphics[width=1.0\textwidth]{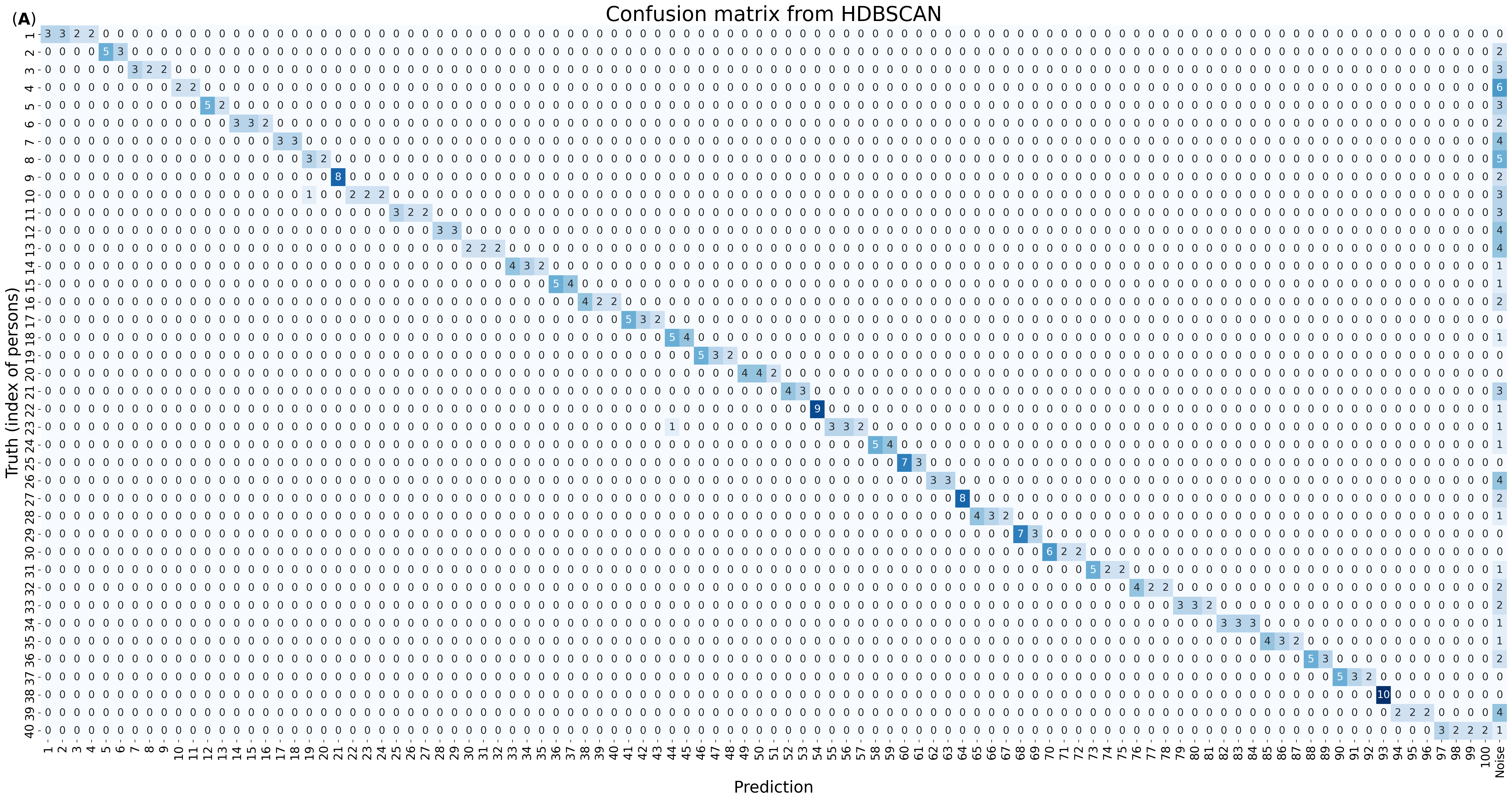}
	\includegraphics[width=0.6\textwidth]{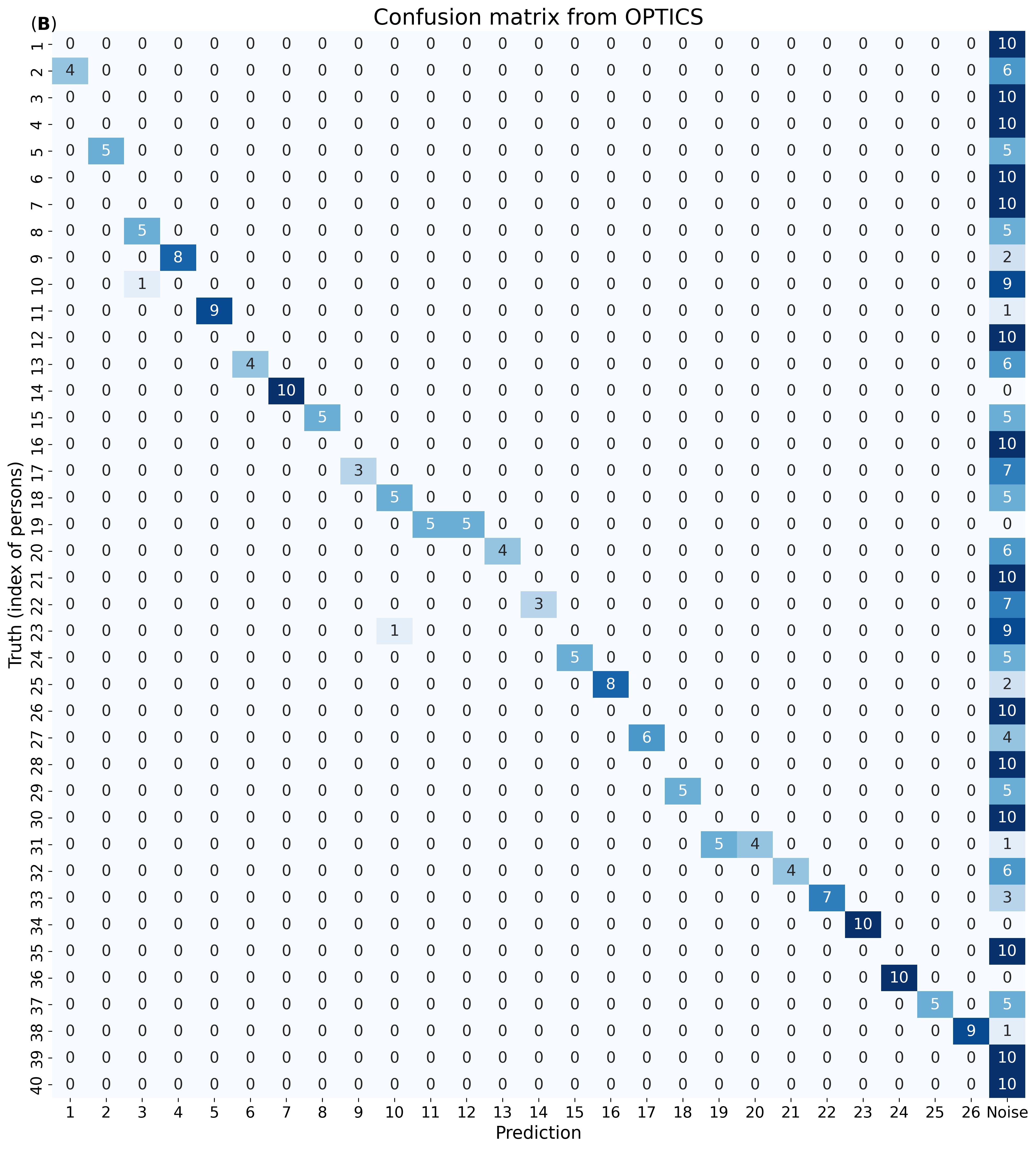}
	\caption{\textbf{Confusion matrices for the clustering of the Olivetti face dataset.} (\textbf{A}) With HDBSCAN with a minimal cluster size of 0.51\% of the data. (\textbf{B}) With OPTICS with a minimal cluster size of 0.51\% of the data.}
	\label{fig:FigureSI-Olivetti-OtherMethods}
\end{figure}

\begin{figure}
	\centering
	\includegraphics[width=0.85\textwidth]{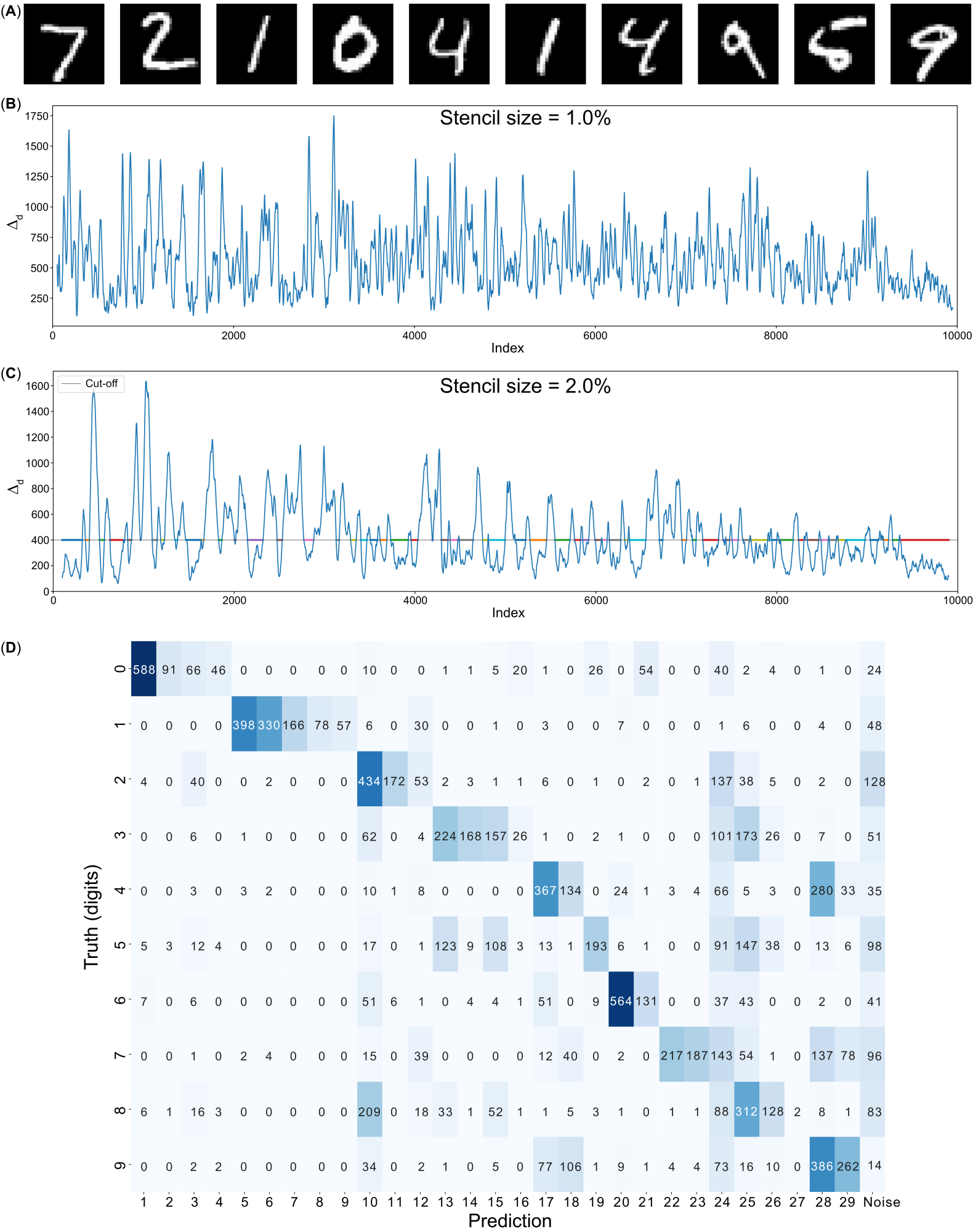}
	\caption{\textbf{Clustering of the MNIST database.} (\textbf{A}) Ten first images from the MNIST database. (\textbf{B}) Plot of $\Delta_d$ with a stencil size of 1.0\%. (\textbf{C}) Plot of $\Delta_d$ with a stencil size of 1.0\%. For panels (\textbf{B}) and (\textbf{C}), the same starting point for the reordering was used (the one that leads to the lowest integral of $\Delta_d$ with a stencil size of 2.0\%). (\textbf{D}) Confusion matrix with no expansion of clusters.}
	\label{fig:FigureSI-MNIST}
\end{figure}

\begin{figure}
	\centering
	\includegraphics[width=1.0\textwidth]{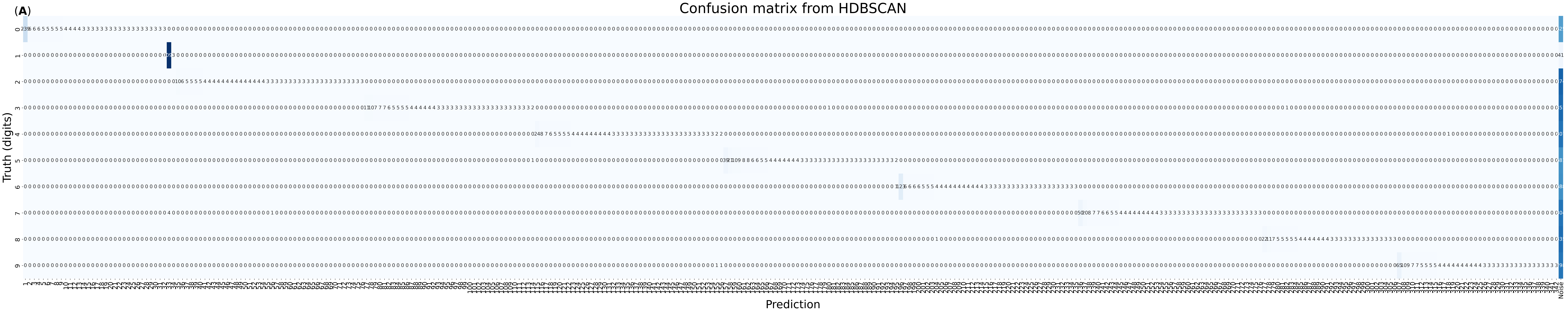}\\
	\includegraphics[width=1.0\textwidth]{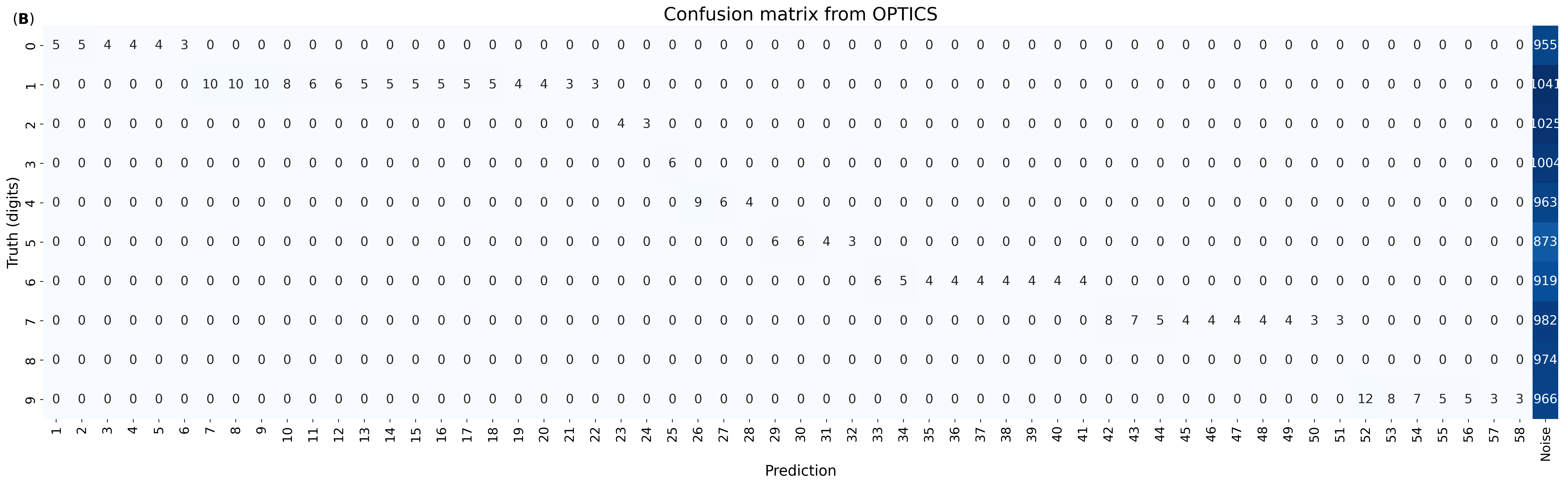}
	\caption{\textbf{Confusion matrices for the clustering of the MNIST database.} (\textbf{A}) With HDBSCAN with a minimal cluster size of 0.13\% of the data. (\textbf{B}) With OPTICS with a minimal cluster size of 0.13\% of the data.}
	\label{fig:FigureSI-MNIST-OtherMethods}
\end{figure}

\begin{figure}
	\centering
    \includegraphics[width=0.9\textwidth,keepaspectratio]{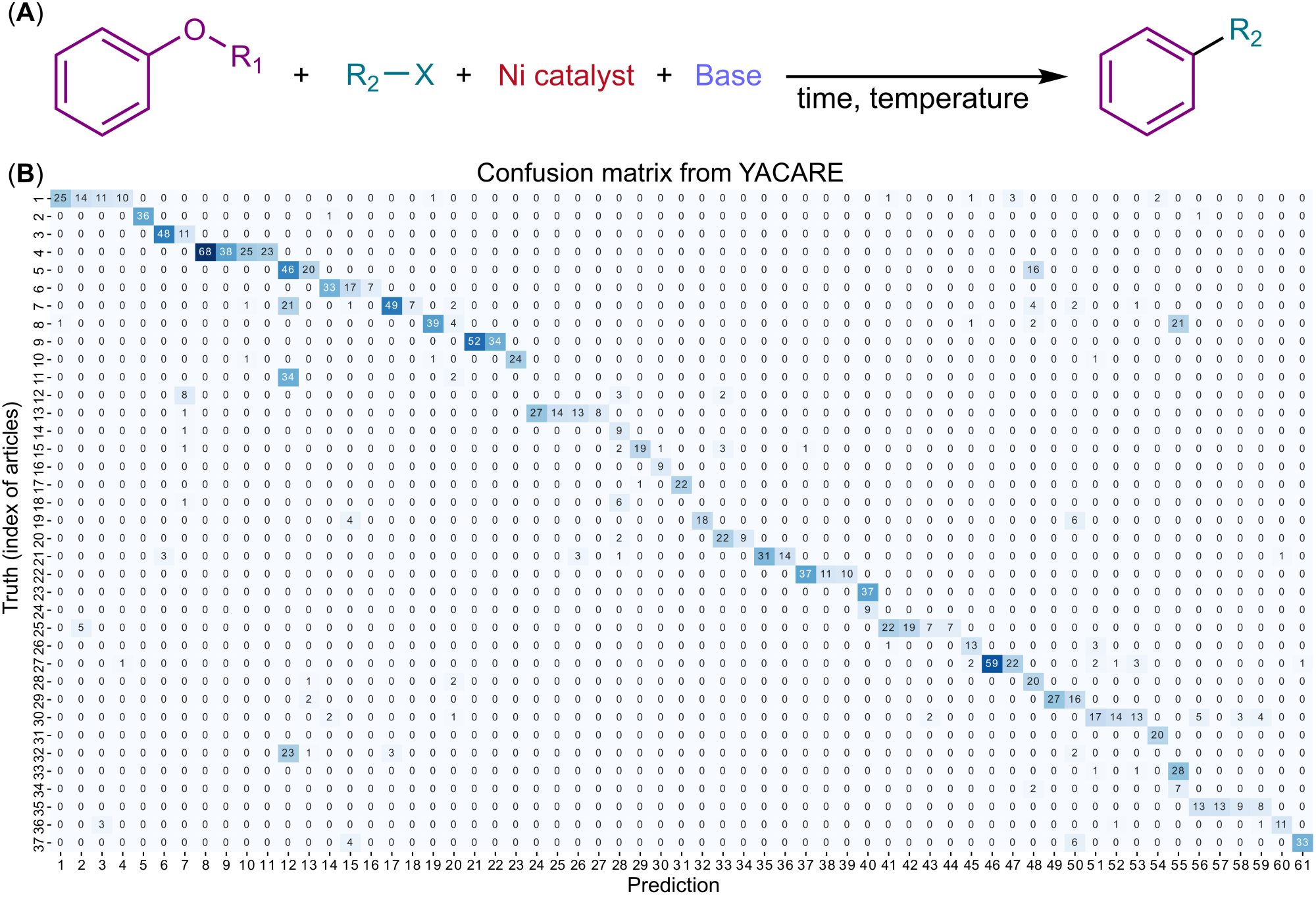}
    \caption{\textbf{Clustering of chemical reactions from the NiCOlit database.} (\textbf{A}) Global scheme for the chemical reactions from the NiCOlit database. (\textbf{B}) Confusion matrices for the clustering of experimental conditions with YACARE, with a minimal cluster size of 0.13\% and a maximal expansion of clusters ($\beta$=$\infty$).}
    \label{fig:FigureSI-NiCOlit}
\end{figure}

\begin{figure}
	\centering
    \includegraphics[width=1.0\textwidth]{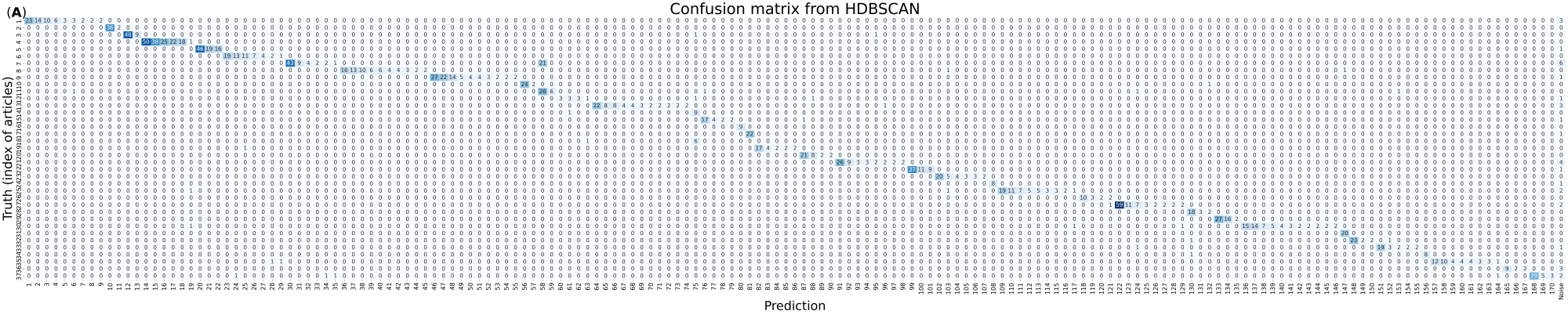} \\
    \includegraphics[width=1.0\textwidth]{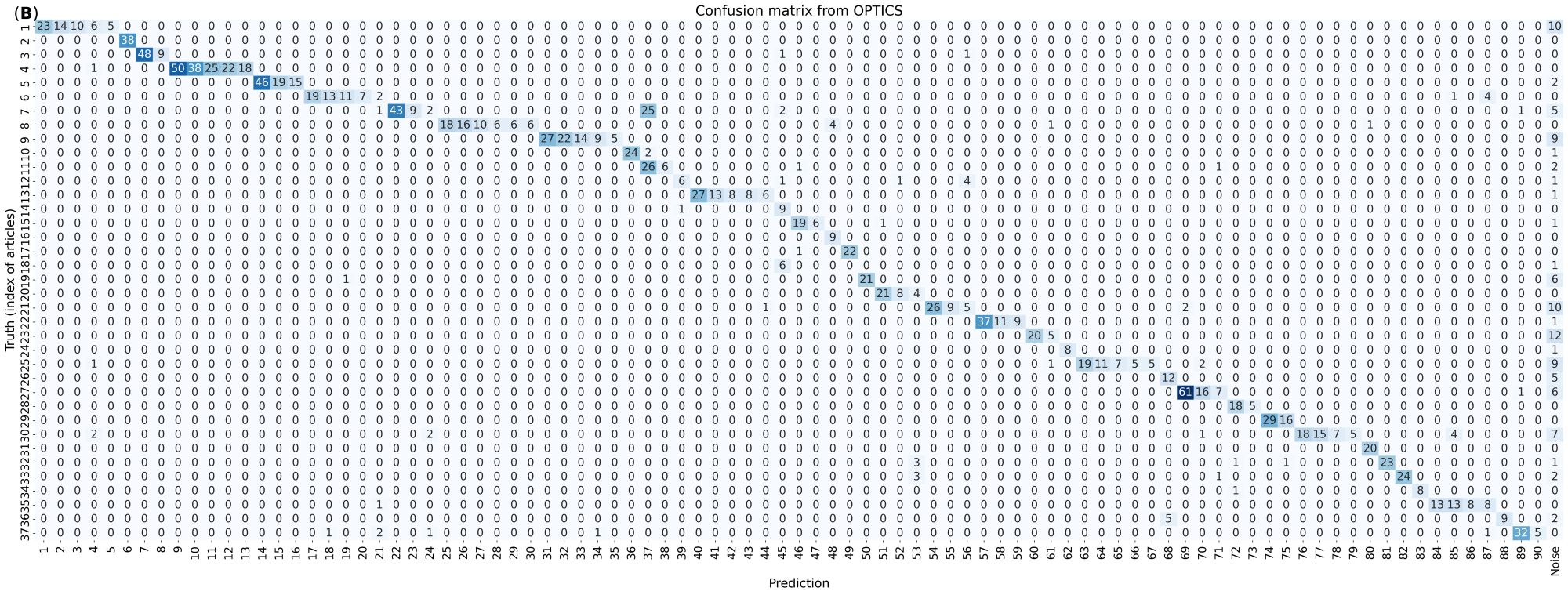} \\
    \includegraphics[width=0.49\textwidth]{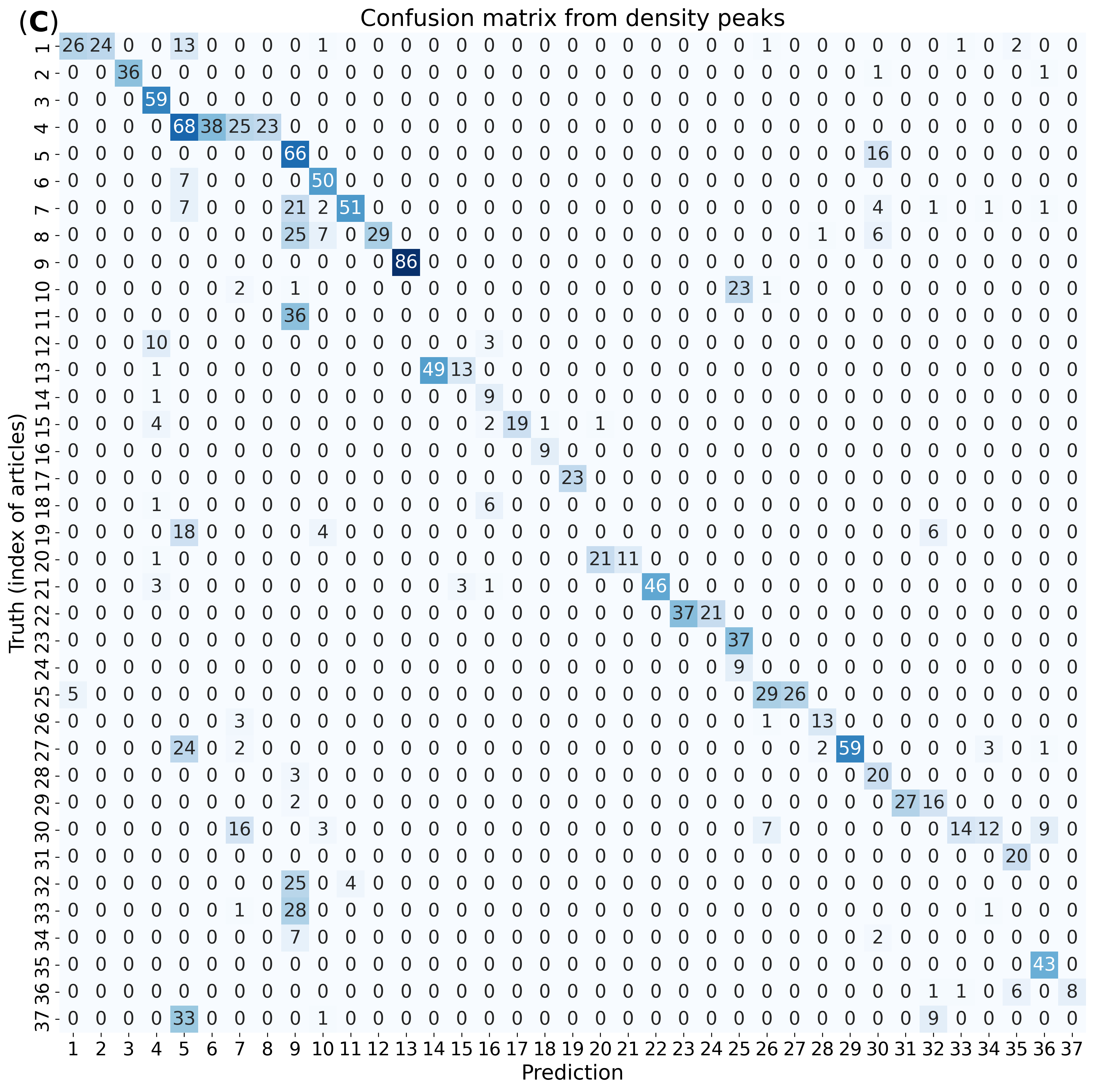}
    \includegraphics[width=0.49\textwidth]{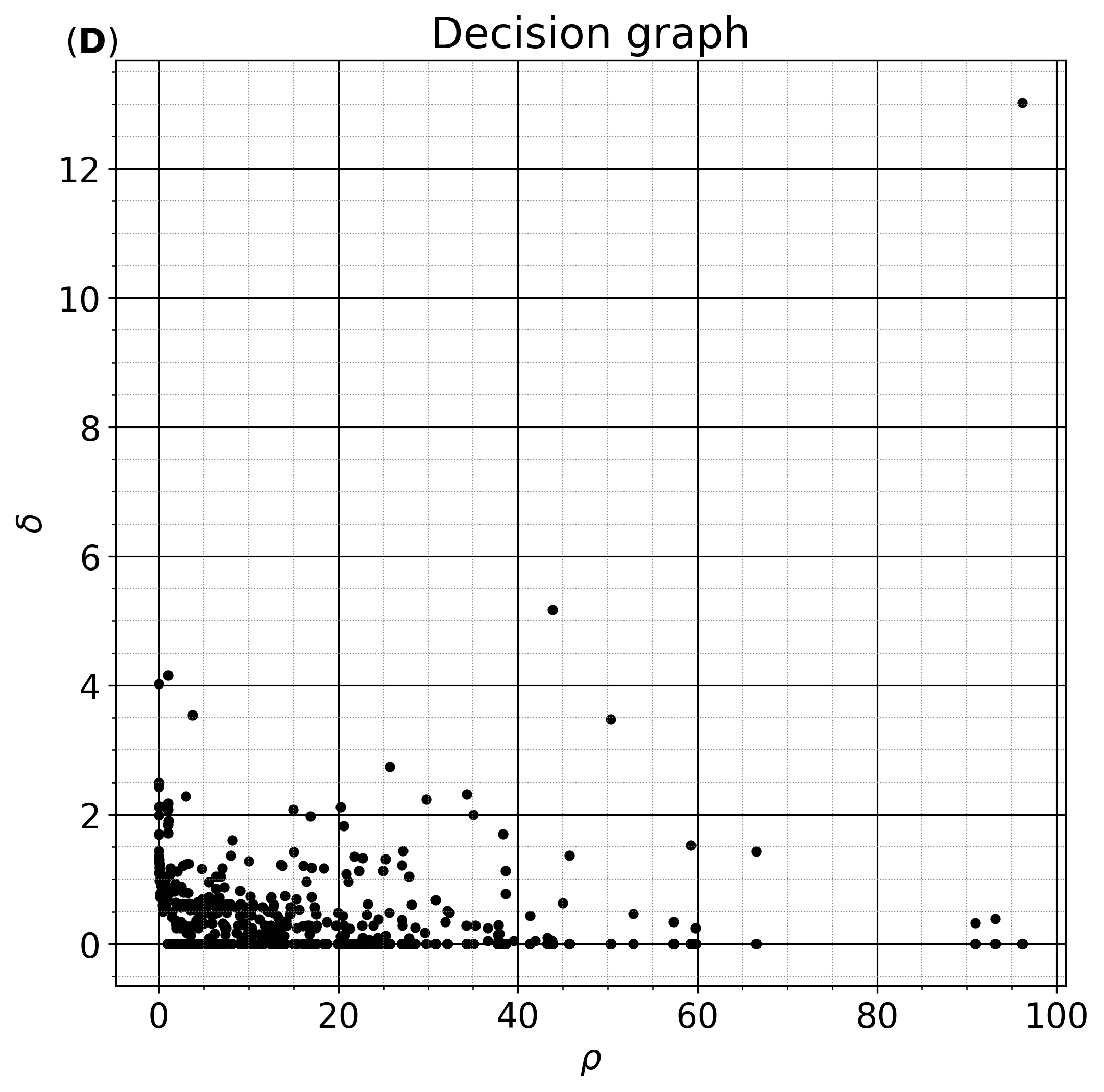} 
    \caption{\textbf{Confusion matrices for the clustering of chemical reactions from the NiCOlit database.} Confusion matrices for the clustering of experimental conditions (\textbf{A}) with HDBSCAN, using a minimal cluster size of 0.13\%, (\textbf{B}) with OPTICS, using a minimal cluster size of 0.13\%, (\textbf{C}) with density peaks with $\rho_{min}$=5.0 and $\delta_{min}$=1.0. (\textbf{D}) Decision graph for density peaks clustering.}
    \label{fig:FigureSI-NiCOlit-OtherMethods}
\end{figure}

\end{document}